\documentclass{article} 
\pdftrailerid{}
\usepackage{iclr2027_conference,times}

\usepackage{amsmath,amsfonts,bm}

\def\eqref#1{equation~\ref{#1}}

\def\1{\bm{1}}

\DeclareMathAlphabet{\mathsfit}{\encodingdefault}{\sfdefault}{m}{sl}
\SetMathAlphabet{\mathsfit}{bold}{\encodingdefault}{\sfdefault}{bx}{n}

\usepackage{algorithm}
\usepackage{float}      
\usepackage{algpseudocode}
\usepackage{hyperref}
\usepackage{url}

\usepackage{graphicx}
\usepackage{booktabs}
\usepackage{array}      
\usepackage{multirow}
\graphicspath{{figs/}}
\newcommand{\method}{LiteEvo}
\usepackage{placeins}   
\usepackage{xcolor}

\title{\method{}: Automated, Cost-Efficient Harness Evolution for Generalization to Unseen Tasks}

\author{%
  Euntae Choi\thanks{Equal contribution.} \\
  Seoul National University \\
  \texttt{euntae.choi175@gmail.com} \\
  \And
  Sumin Song\footnotemark[1] \\
  Seoul National University \\
  \texttt{songsm921@snu.ac.kr} \\
  \And
  Sungjoo Yoo\thanks{Corresponding author.} \\
  Seoul National University \\
  \texttt{sungjoo.yoo@gmail.com} \\
}

\iclrfinalcopy 

\begin{document}

\maketitle
\suppressfloats[t]   

\begin{abstract}
An LLM agent is defined by two things: the weights inside its model and the
harness of components assembled around it. Harnesses are still handcrafted, and HarnessX, which
evolves them automatically, starts each benchmark from a handcrafted harness,
reports gains on the tasks it evolved on, and budgets 100 to 175 million
meta-agent tokens per benchmark. We propose \method{}, a lightweight
harness-evolution algorithm whose tool-free meta-agents mine agent trajectories
for reusable components, curate them into a versioned library, and compose each
round's harness from it, starting every benchmark from the same neutral harness
and never naming the benchmark. Evolving on the graded tasks of five agentic
benchmarks with a frozen Qwen3.5-9B, \method{} lifts pass@2 by 10.5 to
67.7~pp and reaches comparable or higher pass@2 than a reproduction of HarnessX
(71.0 against 67.3 on average) at 13.0$\times$ lower mean API cost. Harnesses
evolved on train tasks keep their gains on unseen test tasks of four
benchmarks, and \method{} also lifts Claude Code with Sonnet~4.6 by 1.2 to
71.4~pp.
\end{abstract}

\section{Introduction}
\label{sec:intro}

Every language-model agent runs inside a harness, the layer that turns the
model's replies into actions and the environment's responses into its next
input. Its prompt, action and observation processing, and loop settings can decide
whether a task is solved: given only the task text, a frozen Qwen3.5-9B solves
none of AppWorld's test tasks, but with a harness that \method{} evolves on
separate training tasks, the same model solves 65.5\% of them within two
attempts (Table~\ref{tab:main_results}).
Changing the harness needs no gradient step and applies to closed models as
well, whereas training the weights of a coding agent with reinforcement
learning takes 512 GPUs for about 32 hours~\citep{swerl}.

\begin{figure}[t]
  \centering
  \includegraphics{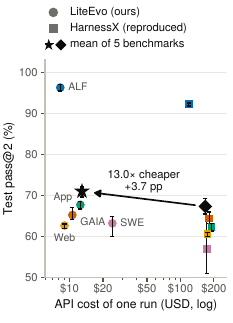}\hfill
  \includegraphics{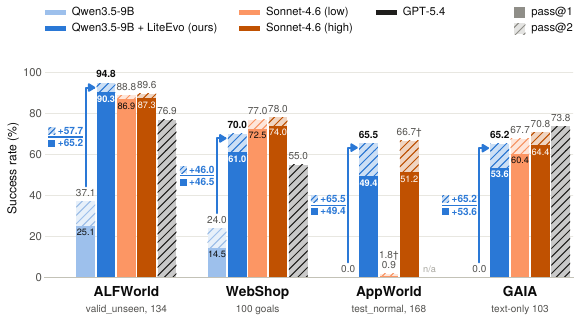}\\[-1pt]
  \makebox[1.55in]{\small (a)}\hfill\makebox[3.85in]{\small (b)}
  \caption{\method{} reaches comparable or higher pass@2 than our reproduction of HarnessX at 13.0$\times$ lower mean API cost and lifts a frozen 9B agent to the level of frontier agents on ALFWorld and AppWorld. (a)~Test pass@2 against the API cost of one evolution run, both methods transductive. Error bars span the three test runs, and the black markers average the five benchmarks. (b)~Qwen3.5-9B given only the task text, and with the harness \method{} evolves (inductive, except on GAIA), against Claude Code with Sonnet~4.6 at low and high effort and GPT-5.4 with the static, unevolved harness that HarnessX reports~\citep{harnessx}, on HarnessX's own task sample for WebShop. Solid bars give pass@1 and hatched bars pass@2. \dag:~Sonnet~4.6 is told how to finish a task; the agent under \method{} never is.}
  \label{fig:teaser}
\end{figure}

Recent work automates harness design with a frontier model that acts as a
meta-agent, reading the agent's traces and rewriting its
harness~\citep{harnessx,metaharness,autoharness}. HarnessX~\citep{harnessx}
applies this idea to five agentic benchmarks, but at a high price, with a
budget of 100 to 175 million meta-agent tokens per benchmark, and with human
knowledge built in: it starts each benchmark from a handcrafted harness, so
every new environment needs new human effort, and the gains it reports mix what
evolution found with what the seed already knew. Its paper also reports gains
at the best round on the tasks it evolved on and does not test whether an
evolved harness helps on tasks it has not seen.

Other systems share some of these limits. Methods in which a model designs
agents automatically report search costs of hundreds to tens of thousands of
dollars per run~\citep{adas,sica,dgm}, and each of the closest harness-evolution
methods tells its optimizer about the benchmark in some form
(Table~\ref{tab:related}): Meta-Harness~\citep{metaharness} guides its
proposer with a hand-written skill for each domain, AutoHarness~\citep{autoharness}
gives its refiner the game's name and description, and
HarnessCompass~\citep{harnesscompass} puts software-repair hints into its meta
prompt. Three recent methods start from a minimal harness or score on
held-out tasks, but each covers at most two of our five
benchmarks~\citep{harnesscompass,selfharness,betterharness}. It therefore
remains open whether a cheap evolution loop that knows nothing about the
benchmark, neither through its starting harness nor through its instructions,
can produce harnesses that help on unseen tasks across agentic benchmarks as
different as ours.

\begin{figure}[t]
  \centering
  \includegraphics[width=\textwidth]{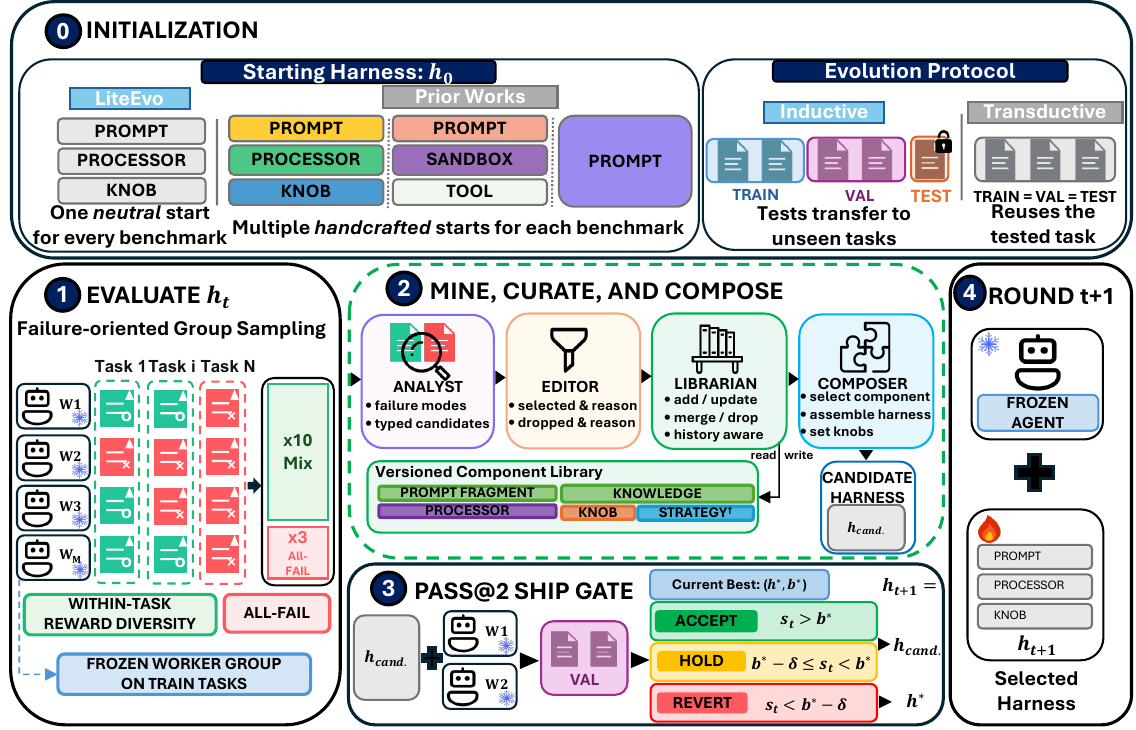}
  \caption{Overview of \method{}. (0)~Every benchmark starts from the same neutral harness $h_0$. (1)~Frozen workers run the current harness $h_t$ on the train tasks, and failure-oriented group sampling keeps ten mixed-outcome and three all-fail task groups as evidence. (2)~Four meta-agents mine failure modes, curate a versioned component library (\dag{} marks the strategies, which advise the composer and never reach the agent) and compose a candidate. (3)~A pass@2 gate on the val tasks accepts, holds or reverts the candidate. (4)~The frozen agent runs the selected harness $h_{t+1}$ in the next round.}
  \label{fig:method}
\end{figure}

We address this question with \method{}, a harness-evolution loop designed for
low API cost and a benchmark-blind start. Every benchmark starts from the same
neutral harness, a 788-character description of the turn protocol with no
benchmark-specific content, and the four meta-agents are never told which
benchmark they work on: a filter masks names and identifiers in everything they
read, and their instructions forbid using recalled knowledge of a benchmark.
Apart from this harness and the benchmark interfaces, the only parts we design
by hand are the loop and the meta-agents' instructions, so the
benchmark-specific content of a returned harness is learned from the agent's
own trajectories. In each round, a single analyst reads a failure-oriented
sample of these trajectories:
the tasks that some attempts solve and others fail, and the lowest-scoring
tasks that no attempt solves. Three more meta-agents turn its diagnosis into
edits of a versioned component library and a new harness, for about five tool-free API calls per round in all; a pass@2 gate
then decides whether to keep that harness (Figure~\ref{fig:method}). We
evaluate \method{} both transductively, as HarnessX does, and inductively, on
test tasks the loop never sees. Our contributions are as follows:

\begin{itemize}  \item \method{}, a generic harness-evolution method that needs no
  benchmark-specific seed: it starts every benchmark from the same neutral
  harness, never tells its four tool-free meta-agents which benchmark they work
  on, and evolves the harness through failure-oriented group sampling, a
  versioned library of typed components and a pass@2 gate.
  \item Comparable or higher pass@2 for a small share of the API cost: with a frozen
  Qwen3.5-9B on ALFWorld, WebShop, AppWorld, GAIA and SWE-bench Verified,
  \method{} reaches a mean pass@2 of 71.0 against 67.3 for a same-stack
  reproduction of HarnessX when both evolve on the graded tasks, at a
  13.0$\times$ lower mean API cost (Figure~\ref{fig:teaser}).
  \item Generalization beyond the evolution tasks: harnesses that \method{}
  evolves on train tasks and selects on val tasks stay within 2.2~pp of the
  transductive ones on unseen test tasks of ALFWorld and AppWorld, exceed them
  on WebShop and SWE-bench Verified (70.0 against 62.7 and 66.7 against 63.2), and reach 48.6 pass@2 on AppWorld tasks that
  need apps unseen in training, where the neutral harness solves none; transfer
  to an unseen benchmark remains open.
\end{itemize}

\section{Preliminaries}
\label{sec:prelim}

\paragraph{Tasks and the Frozen Agent.}
A benchmark provides a set of tasks $\mathcal{T}$, each an interactive
environment whose first observation states the task and which answers every
action with a new observation. When an episode ends, the benchmark's own
evaluator returns a success flag $y\in\{0,1\}$ and a partial credit
$f\in[0,1]$. The agent, a frozen language model with fixed decoding, acts
through a fixed runner: a multi-turn chat loop that passes the action in each
reply to the environment until the episode ends or a per-benchmark turn cap is
reached.

\paragraph{Harness.}
A harness $h=(p,\Phi,\kappa)$ is everything that may change between the
frozen model and the environment. Here $p$ is the system prompt, and $\Phi$ is a set
of \emph{processors}, short programs that the runner calls at fixed points of a
turn. A processor can add a one-turn message before the model call, rewrite
the action after it is extracted, or rewrite the observation after the
environment step. The
last part, $\kappa$, sets five runner \emph{knobs}, which control the reply
protocol, the listing of available actions, an observation length cap, a
last-line action fallback and a display of the remaining turns. The
environment, the turn cap, the decoding parameters and the model lie outside
the harness and never change.

\paragraph{Metric and Regimes.}
We measure a harness by $\mathrm{pass@}k(h;\mathcal{T})$, the probability that
at least one of $k$ independent attempts at a task succeeds, averaged over a
task set $\mathcal{T}$ and estimated with the unbiased estimator of
\citet{chen2021codex}. The tasks fall
into three sets: (i)~\emph{train} tasks $\mathcal{T}_{\mathrm{train}}$, whose
trajectories a method may read, (ii)~\emph{val} tasks
$\mathcal{T}_{\mathrm{val}}$, used to choose between harnesses, and
(iii)~\emph{test} tasks $\mathcal{T}_{\mathrm{test}}$, used to grade the
harness a method returns. The \emph{inductive} regime keeps the three sets
disjoint, so that the test tasks stay unseen during evolution, whereas the
\emph{transductive} regime uses one task set for all three, as HarnessX does
for all its results~\citep{harnessx}. As a floor, the \emph{bare} agent
receives only the benchmark's own task text, with no system prompt.

\section{\method}
\label{sec:method}

\method{} is a fully automated harness-evolution algorithm: given a frozen agent
and a benchmark, it improves the agent's harness over $R$ rounds, starting
from a neutral harness $h_0$ that is the same for every benchmark and using no
human input written for the benchmark (Figure~\ref{fig:method},
Algorithm~\ref{alg:evolve}). Round $t$ runs the current harness $h_t$ with
$M$ frozen workers on the train tasks. Four meta-agents then turn the resulting
trajectories into a candidate $h_{\mathrm{cand}}$: the \emph{analyst}
diagnoses failure modes and proposes components, the \emph{editor} selects and
checks them, the \emph{librarian} records them in a versioned library $\mathcal{L}$ of
components, and the \emph{composer} assembles the candidate from
$\mathcal{L}$, while a journal $\mathcal{J}$ keeps each round's scores,
decisions and failure modes for later rounds. A pass@2 gate on the val tasks
then compares the candidate's score $s_t$ with the score $b^\star$ of the best
harness so far, $h^\star$, and picks the harness $h_{t+1}$ of the next round, either $h_{\mathrm{cand}}$ or $h^\star$.
Round~0 thus runs the neutral harness, and each of the $R-1$ evolution rounds
that follow runs the harness that the gate picked in the round before. In the inductive regime the train and val rollouts come from
disjoint task sets, whereas in the transductive regime both come from the test
tasks and the gate reuses the train rollouts.

The meta-agents read what the agent did and how its episodes ended but never
act in an environment: each of their calls is a single request without tools
that returns structured output. Apart from $h_0$ and the runner that connects
the agent to each benchmark, the only parts we design by hand are the loop and
the meta-agents' instructions, neither of which names or describes any specific
benchmark; Appendix~\ref{app:method} lists all human inputs.

\begin{algorithm}[t]
\caption{One run of \method{} with $R$ rounds. Round~$t$ runs $h_t$: round~0 runs the neutral harness, and the loop picks $h_1,\dots,h_{R-1}$ for the $R-1$ evolution rounds. In the transductive regime, the train and val tasks are the same, and $s_t$ is computed from the train rollouts.}
\label{alg:evolve}
\begin{algorithmic}[1]
\State $\mathcal{L}\gets$ three hand-written strategies;\quad $\mathcal{J}\gets\emptyset$ \Comment{library, journal}
\State $(h^\star,b^\star)\gets\big(h_0,\ \mathrm{pass@2}(h_0;\mathcal{T}_{\mathrm{val}})\big)$ \Comment{best harness and its score}
\For{$t=0,\dots,R-2$} \Comment{picks $h_{t+1}$}
  \State $\mathcal{G}\gets$ one group of $M$ rollouts of $h_t$ per task in $\mathcal{T}_{\mathrm{train}}$ \Comment{Section~\ref{sec:method:sampling}}
  \State $A\gets\textsc{Analyst}\big(\textsc{GroupSample}(\mathcal{G}),\,h_t\big)$ \Comment{Algorithm~\ref{alg:groupsample}}
  \State $E\gets\textsc{Editor}(A,\mathcal{J})$;\quad $\mathcal{L}\gets\textsc{Librarian}(\mathcal{L},E)$ \Comment{Section~\ref{sec:method:curate}}
  \State $h_{\mathrm{cand}}\gets\textsc{Composer}(\mathcal{L},E,\mathcal{J},h^\star)$
  \State $s_t\gets\mathrm{pass@2}(h_{\mathrm{cand}};\mathcal{T}_{\mathrm{val}})$, from $M$ rollouts per val task
  \If{$s_t>b^\star$} \Comment{accept}
    \State $h_{t+1}\gets h_{\mathrm{cand}}$;\quad $(h^\star,b^\star)\gets(h_{\mathrm{cand}},s_t)$
  \ElsIf{$s_t<b^\star-\delta$} \Comment{revert}
    \State $h_{t+1}\gets h^\star$
  \Else \Comment{hold}
    \State $h_{t+1}\gets h_{\mathrm{cand}}$
  \EndIf
  \State append round $t$ (scores, decision, failure modes) to $\mathcal{J}$
\EndFor
\State \Return $h^\star$
\end{algorithmic}
\end{algorithm}

\subsection{A neutral start and benchmark-blind meta-agents}
\label{sec:method:neutral}

\method{} starts all five benchmarks from one \emph{neutral harness} $h_0$,
whose 788-character prompt $p$ states only the protocol of an episode: the
agent reads an observation, replies with one action inside action tags, and
has a limited number of turns. Where actions are code, the reply format allows
a multi-line action. Since $h_0$ has no processors and uses default knob
values, what a returned harness knows about its benchmark comes from
evolution, apart from generic advice in the meta-agents' instructions.
HarnessX, by contrast, starts each benchmark from a handcrafted
harness~\citep{harnessx}; its GAIA prompt alone holds about 11{,}000
characters of hand-written search recipes, several of which name specific
websites.

The meta-agents are blind to the benchmark as well. Before a meta-agent reads a
trajectory, a filter removes task identifiers, split names and URLs, masks
benchmark and dataset names, and shows the turn limit only as short or long, a
label that at least two benchmarks share. The meta-agents' instructions forbid
relying on recalled knowledge of a benchmark, even if they recognize it, and
Appendix~\ref{app:method} details how both rules are enforced.

\subsection{Failure-oriented group sampling}
\label{sec:method:sampling}

Each round evaluates the current harness with $M$ frozen \emph{workers}, each
of which runs $h_t$ once on every train task, so that every task yields $M$
trajectories that form its \emph{group}. The groups are the input of the
analyst, the first meta-agent, but it does not read all of them. A group is
\emph{mixed} when some but not all workers solve the task, and \emph{all-fail}
when none does; with \emph{failure-oriented group sampling}, the analyst reads
the ten mixed groups with the highest mean partial credit and the three
all-fail groups with the lowest. Each comes with all $M$ trajectories and their
outcomes, together with outcome statistics over the whole round and the prompt
$p$ that the agent ran (Algorithm~\ref{alg:groupsample}). The two kinds serve different ends. A success and a failure in one mixed
group share the task and the harness and differ only by sampling, so the step
that separates them is one that the harness leaves to chance, and fixing it
makes the harness more reliable (exploitation). An all-fail group instead shows
a task that the harness does not yet make solvable, and asks for the insight
that would (exploration). Without both kinds, evolution stalls where successes
are rare and hard to reach, as on AppWorld: reading only all-success or only
all-fail groups ends there at 16.9 and 12.3 pass@2, against 67.7 with both
(Appendix~\ref{app:sampling}). The design echoes group-relative reinforcement
learning, which learns only from groups whose rewards
differ~\citep{grpo,dapo}, and Training-Free GRPO, which turns such
within-group contrasts into prompt-level experience~\citep{tfgrpo}; unlike
them, \method{} also keeps all-fail groups, as targets for exploration.

\subsection{The meta-agent stage}
\label{sec:method:curate}

The library $\mathcal{L}$ stores typed components that fill the three parts of
a harness $h=(p,\Phi,\kappa)$. Prompt fragments, which are instructions or
worked examples, and knowledge components, which record facts about the
environment seen in trajectories, go into the prompt $p$; processors go into
$\Phi$, and knob components hold settings for $\kappa$. A fifth kind,
strategies, advises the composer and never reaches the agent. Initially the
library is empty except for three hand-written strategies, generic composing
rules such as leaving the reply format to the runner
(Appendix~\ref{app:method}). The edits of a round produce a new library
version, and components record when they were selected and how often they
fired.
On AppWorld, for instance, the returned harness holds a worked login example
as a fragment and the login API's credential fields as knowledge, and one of
its processors restates the login steps after two authorization errors.

\paragraph{Analyst.}
The analyst first compares the trajectories within each sampled group and
only then generalizes across groups. It returns a list of failure modes, giving
for each one its share of the failures, the evidence and the mechanism behind
it. For each mode, it also proposes candidate components of a given kind,
written out in full. Its instructions ask for fixes that the harness can carry
out, such as a worked example or a processor, rather than general advice to the
agent.

\paragraph{Editor.}
The editor merges, rewrites or drops the analyst's candidates with a reason
for each decision, reading a journal $\mathcal{J}$ of earlier rounds (their
gate scores, decisions, failure modes and rejected components) so that it does
not repeat a change that already failed. Every processor it keeps is compiled
and replayed on 24 episodes of the round, and one that does not compile,
raises an error or never fires is dropped; a separate review call then reads
the outputs of each remaining processor and drops those that contradict its
stated purpose.

\paragraph{Librarian.}
The librarian applies the editor's selection to the library: it sees all
components with their usage statistics and returns operations that add a
component, update or merge existing ones, or drop one. Each operation is
validated again before it is applied: a processor must compile, a knob setting
must parse, and no body may name a benchmark. The library is never rolled
back, and a component therefore survives even when the harness that first used
it is reverted.

\paragraph{Composer.}
The composer writes the candidate harness from the library: it selects
components, writes a free-text frame that opens the prompt, and sets the five
knobs. It sees the library's index and bodies, the journal, the selection and
knobs of the best harness $h^\star$, the editor's picks, and a short excerpt of
failed trajectories. A fixed assembly then builds
$h_{\mathrm{cand}}=(p,\Phi,\kappa)$: $p$ joins the frame, the selected prompt
fragments and knowledge, and the runner's reply-format block; $\Phi$ holds the
selected processors; and $\kappa$ takes the composer's knob values. The
composer builds each candidate afresh from the library instead of editing the
previous harness.

\subsection{The pass@2 gate}
\label{sec:method:gate}

The gate keeps the best harness so far, $h^\star$, and its val score $b^\star$,
initially $h_0$ and its own score. It scores a candidate by
$s_t=\mathrm{pass@2}(h_{\mathrm{cand}};\mathcal{T}_{\mathrm{val}})$, the
metric we report, estimated from $M$ rollouts per val task, and picks the
harness of the next round:
\begin{equation}
  h_{t+1}=
  \begin{cases}
    h_{\mathrm{cand}}, \text{ and } (h^\star,b^\star)\leftarrow(h_{\mathrm{cand}},s_t)
      & \text{if } s_t>b^\star \quad\text{(accept)},\\
    h^\star & \text{if } s_t<b^\star-\delta \quad\text{(revert)},\\
    h_{\mathrm{cand}} & \text{otherwise} \quad\text{(hold)}.
  \end{cases}
  \label{eq:gate}
\end{equation}
A hold keeps searching from the candidate without making it the best harness.
The margin $\delta$ is the largest of three quantities
(Appendix~\ref{app:method}): (i)~7~pp, or half of $b^\star$ if that
is smaller, (ii)~a noise band estimated from how far the workers' scores
spread within the rounds so far, and (iii)~the spread of the best harness's
repeated measurements. A candidate whose replies fail to parse more than 5\% of
the time is never accepted. After $R$ rounds, \method{} returns $h^\star$, and
a result's best round $t$ means that the returned harness is $h_t$, first run
in round $t$. In the loop itself, a candidate's val rollouts run at the
start of the next round together with its train rollouts, so the meta-agents
have already composed from them when the gate decides; a revert discards that
composition, while the library keeps its edits.

\subsection{Why \method{} is cheap}
\label{sec:method:cost}

Three design choices keep the API cost of \method{} low. First, every
meta-agent call is a single tool-free request on an input that the loop
assembles, so that a round makes about five calls, one per meta-agent plus the
editor's processor review, which Figure~\ref{fig:method} and
Algorithm~\ref{alg:evolve} omit for brevity. An agentic meta-agent such as
HarnessX's instead works through dozens of tool-using turns per round and
re-sends its growing context at each of them~\citep{harnessx}. Second, only
the analyst reads full trajectories, and only a fixed-size sample of them, on
the cheaper of the two models, while the other meta-agents read far shorter
inputs, such as its structured output, the journal and the library. Third, the
library persists across rounds, so that a round adds or revises a few
components instead of re-deriving the harness from the trajectories.
Section~\ref{sec:results} measures the resulting cost.

\section{Evaluation}
\label{sec:results}

\subsection{Experimental setup}
\label{sec:results:setup}

\paragraph{Models and Benchmarks.}
We evaluate on ALFWorld~\citep{alfworld}, WebShop~\citep{webshop},
AppWorld~\citep{appworld}, GAIA~\citep{gaia} and SWE-bench
Verified~\citep{swebench}, which ask for household tasks in a text world,
product searches, everyday tasks through the APIs of nine apps, web research
with code, and repository-level bug fixes. The frozen agent is Qwen3.5-9B in
thinking mode, served locally, and the analyst runs on Claude Sonnet~5 and the
other three meta-agents on Claude Opus~4.6.

\paragraph{Implementation Details.}
Each run lasts $R=8$ rounds, round~0 with the neutral harness and seven
evolution rounds, with $M=4$ workers ($M=2$ on SWE-bench Verified). Unless a caption says otherwise, we grade a returned
harness in three test runs of two attempts per task and report the mean pass@1
and pass@2.
The API cost of a run counts the meta-agents' tokens in the calls that produce
$h_1$ to $h_7$, at Opus~4.6's list price and Sonnet~5's introductory price.
We leave out the cost of running the
agent, which runs on local GPUs, for all methods: every harness-evolution
method pays it, and comparing it fairly would need one common setup for running
the agent, which was not yet fully reproducible at the time of submission.

\paragraph{Baselines.}
We compare with the bare agent and with HarnessX~\citep{harnessx}, which we
reproduce from its released code with the same agent and tasks, using
Opus~4.6 as its meta-agent for eight rounds. As in its paper, HarnessX evolves
on the tasks it is graded on, and we report its best round; we wrote its seeds
for ALFWorld, WebShop and AppWorld, which the release lacks.
On GAIA, whose agent searches the live web, \method{}'s web tools block known
copies of the benchmark's answers. For details of the setup not mentioned
here, see Appendix~\ref{app:benchmarks}.

\subsection{Main results}
\label{sec:results:main}
\begin{table*}[t]
    \centering
    \caption{Test pass@2 (\%) of Qwen3.5-9B, averaged over three test runs. \textit{Bare} gives the agent only the task text. \textit{Initial} is the starting harness, a benchmark-specific seed for HarnessX (+Seed) and the neutral harness for \method{} (+Neutral), and \textit{Evolved} is the returned harness $h_t$, first run in round~$t$ (\textit{Best Round}). $\Delta$ is Evolved minus Initial in pp, computed before rounding.}
    \label{tab:main_results}
    \small
    \setlength{\tabcolsep}{3.4pt}
    \renewcommand{\arraystretch}{1.08}
    \begin{tabular}{lllcccccc}
        \toprule
        Benchmark & Mode & Method & Initialization
        & \multicolumn{2}{c}{pass@2 (\%)} & $\Delta$ & \shortstack{Best\\Round} & \shortstack{API cost\\(\$)} \\
        \cmidrule(lr){5-6}
        & & & & Initial & Evolved & & & \\
        \midrule
        \multirow{4}{*}{ALFWorld}
            & Bare                          & --       & Task only  & 37.1 &      & --    & -- & --        \\
        \cmidrule(lr){2-9}
            & \multirow{2}{*}{Transductive} & HarnessX & +Seed      & 64.2 & 92.3 & +28.1 & 6  & 118.47 \\
            &                               & \method{}  & +Neutral   & 60.9 & 96.3 & +35.3 & 5  & 8.24  \\
        \cmidrule(lr){2-9}
            & Inductive                     & \method{}  & +Neutral   & 60.9 & 94.8 & +33.8 & 4  & 9.17  \\
        \midrule
        \multirow{4}{*}{WebShop}
            & Bare                          & --       & Task only  & 24.0 &      & --    & -- & --        \\
        \cmidrule(lr){2-9}
            & \multirow{2}{*}{Transductive} & HarnessX & +Seed      & 45.3 & 60.7 & +15.3 & 6  & 175.11 \\
            &                               & \method{}  & +Neutral   & 10.0 & 62.7 & +52.7 & 5  & 9.00  \\
        \cmidrule(lr){2-9}
            & Inductive                     & \method{}  & +Neutral   & 10.0 & 70.0 & +60.0 & 3  & 9.89  \\
        \midrule
        \multirow{4}{*}{AppWorld}
            & Bare                          & --       & Task only  & 0.0  &      & --    & -- & --        \\
        \cmidrule(lr){2-9}
            & \multirow{2}{*}{Transductive} & HarnessX & +Seed      & 48.6 & 62.3 & +13.7 & 6  & 190.32 \\
            &                               & \method{}  & +Neutral   & 0.0  & 67.7 & +67.7 & 6  & 12.51  \\
        \cmidrule(lr){2-9}
            & Inductive                     & \method{}  & +Neutral   & 0.0  & 65.5 & +65.5 & 7  & 14.56  \\
        \midrule
        \multirow{3}{*}{GAIA}
            & Bare                          & --       & Task only  & 0.0  &      & --    & -- & --        \\
        \cmidrule(lr){2-9}
            & \multirow{2}{*}{Transductive} & HarnessX & +Seed      & 56.0 & 64.4 & +8.4  & 2  & 181.89 \\
            &                               & \method{}  & +Neutral   & 48.5 & 65.2 & +16.7 & 7  & 10.59  \\
        \midrule
        \multirow{4}{*}{SWE-bench Verified}
            & Bare                          & --       & Task only  & 16.4 &      & --    & -- & --        \\
        \cmidrule(lr){2-9}
            & \multirow{2}{*}{Transductive} & HarnessX & +Seed      & 53.9 & 57.0 & +3.0  & 1  & 172.22 \\
            &                               & \method{}  & +Neutral   & 52.7 & 63.2 & +10.5 & 6  & 24.30  \\
        \cmidrule(lr){2-9}
            & Inductive                     & \method{}  & +Neutral   & 52.7 & 66.7 & +13.9 & 4  & 28.50  \\
        \bottomrule
    \end{tabular}
\end{table*}

\paragraph{Main Comparison.}
Table~\ref{tab:main_results} compares \method{} with HarnessX, and
Figure~\ref{fig:teaser}a plots each run's pass@2 against its API cost. In the
transductive regime, where both methods evolve on the tasks they are graded
on, \method{} reaches a mean pass@2 of 71.0 over the five benchmarks, compared
with 67.3 for HarnessX. It spends \$8.24 to \$24.30 per run where HarnessX
spends \$118.47 to \$190.32, a 13.0$\times$ lower mean API cost. It does so
from a weaker start on all five benchmarks: the neutral harness solves no
AppWorld task, whereas the handcrafted seed reaches 48.6, and it scores 10.0 on
WebShop, where the seed reaches 45.3. Evolution closes both gaps within eight
rounds, adding 67.7 and 52.7~pp of pass@2.
The neutral harness is not always the strongest trivial start: the bare agent
scores 24.0 on WebShop, and a one-line format hint reaches 61.2 on SWE-bench
Verified (Table~\ref{tab:qwen9b-prompt-processor-ablation}). Measured from these
stronger starts, \method{}'s transductive gains are 38.7 and 2.0~pp.
\providecommand{\method}{LiteEvo}
\begin{table}[t]
    \centering
    \caption{The returned harness (\textit{Full}) split into its system prompt and processors (Qwen3.5-9B, transductive): \textit{+Prompt} and \textit{+Processor} add one part to \textit{Bare + format}, the bare agent with a minimal action-format hint. P@1 and P@2 are pass@1 and pass@2 in \%, averaged over three test runs (one for Bare on WebShop, GAIA and SWE-bench Verified).}
    \label{tab:qwen9b-prompt-processor-ablation}
    \small
    \setlength{\tabcolsep}{5pt}
    \renewcommand{\arraystretch}{1.08}
    \begin{tabular}{lcccccccccc}
      \toprule
      & \multicolumn{2}{c}{ALFWorld} & \multicolumn{2}{c}{WebShop} & \multicolumn{2}{c}{AppWorld}
      & \multicolumn{2}{c}{GAIA} & \multicolumn{2}{c}{\shortstack{SWE-bench\\Verified}} \\
      \cmidrule(lr){2-3} \cmidrule(lr){4-5} \cmidrule(lr){6-7} \cmidrule(lr){8-9} \cmidrule(lr){10-11}
      Harness & P@1 & P@2 & P@1 & P@2 & P@1 & P@2 & P@1 & P@2 & P@1 & P@2 \\
      \midrule
      Bare          & 25.1 & 37.1 & 14.5 & 24.0 & 0.0  & 0.0  & 0.0  & 0.0  & 9.1  & 16.4 \\
      Bare + format & 46.8 & 56.7 & 9.3  & 13.8 & 0.0  & 0.0  & 29.1 & 44.3 & 50.9 & 61.2 \\
      \midrule
      +Processor    & 49.3 & 57.7 & 41.3 & 53.0 & 0.0  & 0.0  & 44.6 & 59.2 & 51.0 & 63.1 \\
      +Prompt       & \textbf{92.8} & 95.5 & 51.0 & 59.0 & 46.1 & 63.1 & 41.9 & 49.5 & 49.4 & 58.2 \\
      \midrule
      Full          & 92.6 & \textbf{96.3} & \textbf{54.0} & \textbf{62.7} & \textbf{51.9} & \textbf{67.7}
                    & \textbf{53.6} & \textbf{65.2} & \textbf{54.2} & \textbf{63.2} \\
      \bottomrule
    \end{tabular}
\end{table}

\paragraph{Cost Breakdown.}
As Section~\ref{sec:method:cost} anticipates, HarnessX's meta-agent, an
agentic session with file tools, makes 65 to 130 Opus
calls and consumes 3.3 to 5.6 million input tokens per round in our
reproduction, whereas \method{} makes about five calls and consumes 0.12 to
0.54 million, mostly read by the analyst on the cheaper Sonnet~5.
\begin{table*}[t]
    \centering
    \caption{pass@2 (\%) of Claude Code with Sonnet~4.6 at low effort and its own tools turned off, given only the task (\textit{Initial}) and with the harness that \method{} evolves for it; each evolved cell is a single measurement. The GAIA run blocks known copies of the benchmark's answers. In Figure~\ref{fig:teaser}b, Sonnet~4.6 on AppWorld is also told how to finish a task.}
    \label{tab:claude_code_evolite}
    \small
    \setlength{\tabcolsep}{5pt}
    \renewcommand{\arraystretch}{1.08}
    \begin{tabular}{lllccccc}
        \toprule
        Benchmark & Mode & Method
        & \multicolumn{2}{c}{pass@2 (\%)} & $\Delta$ & \shortstack{Best\\Round} & API cost (\$) \\
        \cmidrule(lr){4-5}
        & & & Initial & Evolved & & & \\
        \midrule
        \multirow{2}{*}{ALFWorld}
            & Transductive & \method{} & 88.8 & 96.3 & +7.5  & 2 & 8.21 \\
            & Inductive    & \method{} & 88.8 & 95.5 & +6.7  & 7 & 12.92 \\
        \midrule
        \multirow{2}{*}{WebShop}
            & Transductive & \method{} & 77.0 & 81.0 & +4.0  & 6 & 9.84 \\
            & Inductive    & \method{} & 77.0 & 81.0 & +4.0  & 5 & 12.57 \\
        \midrule
        \multirow{2}{*}{AppWorld}
            & Transductive & \method{} & 0.0  & 66.1 & +66.1 & 7 & 11.09 \\
            & Inductive    & \method{} & 0.0  & 71.4 & +71.4 & 4 & 12.58 \\
        \midrule
        GAIA
            & Transductive & \method{} & 67.7 & 74.3 & +6.7  & 4 & 9.08 \\
        \midrule
        \multirow{2}{*}{SWE-bench Verified}
            & Transductive & \method{} & 77.0 & 80.0 & +3.0  & 2 & 11.63 \\
            & Inductive    & \method{} & 77.0 & 78.2 & +1.2  & 6 & 11.80 \\
        \bottomrule
    \end{tabular}
\end{table*}

\paragraph{Inductive Transfer.}
The inductive runs of Table~\ref{tab:main_results} evolve on train tasks,
choose between harnesses on val tasks, and are graded on test tasks that the
loop never sees. On ALFWorld and AppWorld they stay within 2.2~pp of the
transductive runs, and on WebShop and SWE-bench Verified they score higher,
with 70.0 against 62.7 and 66.7 against 63.2;
the inductive harnesses also score above HarnessX's transductive ones, although
HarnessX, which reports no held-out evaluation, selected its harnesses on the
test tasks themselves. GAIA has
no inductive row, as its
103 public questions are too few to split three ways. On AppWorld tasks that
need apps unseen in training, our inductive harness reaches 48.6 where the
neutral harness solves none, against 49.2 for our transductive one and 50.5 for
HarnessX's (Appendix~\ref{app:aw-challenge}); a library evolved on other
benchmarks does not yet help an unseen one.


\paragraph{Other Backbones.}
Table~\ref{tab:claude_code_evolite} applies \method{} to Claude
Code~\citep{claudecode} running Sonnet~4.6: the composed system prompt is
appended to Claude Code's own, and the processors act on the actions and
observations exchanged with the environment. \method{} raises pass@2 in all nine benchmark and regime cells at
\$8.21 to \$12.92 per run: by 1.2 to 7.5~pp on the four benchmarks where the
bare agent is already strong, and on AppWorld from 0.0 to 66.1 in the
transductive regime and to 71.4 in the inductive one. On the
smaller Qwen3.5-4B, \method{} adds 6.7 to 49.8~pp to the neutral start (Appendix~\ref{app:4b}).

\subsection{Ablation studies}
\label{sec:results:analysis}
This section examines how the returned harness splits between its prompt and
its processors, how the library grows, and whether a handcrafted start helps;
for more ablations, see Appendix~\ref{app:results}.

\paragraph{Prompt versus Processors.}
Table~\ref{tab:qwen9b-prompt-processor-ablation} splits a returned harness
into its system prompt and its processors. The full harness gives the best
pass@2 on all five benchmarks. The prompt carries most of
the gain on ALFWorld and AppWorld, reaching 95.5 and 63.1 alone compared with
96.3 and 67.7 for the full harness; on GAIA and SWE-bench Verified the
processors do, reaching 59.2 and 63.1 alone compared with 65.2 and 63.2. Since the two parts evolved
together, the split does not isolate their contributions.

\providecommand{\method}{LiteEvo}
\begin{table}[t]
    \centering
    \caption{Library operations and composer selections per round in the five transductive \method{} runs (Qwen3.5-9B), summed over the runs; \textit{Size} (components selectable after the round) and \textit{Picked} (selected for the next harness) are averaged. \textit{Removed} counts components deleted by merges and drops, and \textit{Pick rate} is the pooled ratio of Picked to Size.}
    \label{tab:qwen9b-library-ops-pooled}
    \small
    \setlength{\tabcolsep}{5pt}
    \renewcommand{\arraystretch}{1.08}
    \begin{tabular}{@{}llccccccccc@{}}
      \toprule
      \multicolumn{2}{@{}l}{Round} & 0 & 1 & 2 & 3 & 4 & 5 & 6 & 7 & Total \\
      \midrule
      \multirow{4}{*}{Librarian}
        & Add    & 19 & 11 & 12 & 10 & 6 & 11 & 8 & 8 & 85 \\
        & Update & 0  & 4  & 4  & 2  & 5 & 0  & 3 & 7 & 25 \\
        & Merge  & 0  & 1  & 3  & 2  & 0 & 1  & 0 & 0 & 7  \\
        & Drop   & 0  & 0  & 1  & 0  & 0 & 0  & 0 & 1 & 2  \\
      \midrule
      \multirow{2}{*}{Library}
        & Removed & 0   & 2   & 4   & 2    & 0    & 0    & 0    & 1    & 9  \\
        & Size    & 3.8 & 5.8 & 8.0 & 10.0 & 11.2 & 13.4 & 15.0 & 16.4 & -- \\
      \midrule
      \multirow{2}{*}{Composer}
        & Picked    & 3.8   & 5.8   & 7.8  & 9.6  & 9.8  & 11.4 & 11.2 & 12.4 & -- \\
        & Pick rate (\%) & 100.0 & 100.0 & 97.5 & 96.0 & 87.5 & 85.1 & 74.7 & 75.6 & -- \\
      \bottomrule
    \end{tabular}
\end{table}

\paragraph{Library Growth.}
In the five transductive runs (Table~\ref{tab:qwen9b-library-ops-pooled}),
the librarian mostly adds, with 85 additions against 25 updates, 7 merges and 2
drops, so the library grows from 3.8 components on
average after the first round to 16.4 after the last. The composer uses all
components in the first two rounds and then becomes selective, picking 75.6\%
of them in the last round. Forcing an explicit memory across episodes into
every harness lowered pass@2 (Appendix~\ref{app:memory}).

\providecommand{\method}{LiteEvo}
\begin{table}[t]
    \centering
    \caption{Neutral or handcrafted start (Qwen3.5-9B, transductive, one run each; HarnessX's own seed on GAIA): test P@1 and P@2 (\%), mean of three test runs.}
    \label{tab:qwen9b-init-ablation}
    \small
    \setlength{\tabcolsep}{5pt}
    \renewcommand{\arraystretch}{1.08}
    \begin{tabular}{lcccc}
      \toprule
      & \multicolumn{2}{c}{AppWorld} & \multicolumn{2}{c}{GAIA} \\
      \cmidrule(lr){2-3} \cmidrule(lr){4-5}
      Initial harness & P@1 & P@2 & P@1 & P@2 \\
      \midrule
      Neutral (default) & \textbf{51.9} & \textbf{67.7} & \textbf{53.6} & \textbf{65.2} \\
      HarnessX seed     & 49.1 & 64.9 & 48.4 & 59.9 \\
      \bottomrule
    \end{tabular}
\end{table}

\paragraph{Neutral versus Handcrafted Start.}
Table~\ref{tab:qwen9b-init-ablation} starts \method{} from a handcrafted
HarnessX-style seed instead of the neutral harness. The seeded runs end lower
on both benchmarks, with 64.9 against 67.7 on AppWorld and 59.9 against 65.2 on
GAIA, so the neutral start matches or exceeds a handcrafted one and is our
default; with one run per arm, this does not show that a seed hurts.

\FloatBarrier
\section{Conclusion}
\label{sec:conclusion}
We introduced \method{}, which evolves a harness from the same neutral start
on every benchmark with four tool-free meta-agents, a versioned component
library and a pass@2 gate. With a frozen Qwen3.5-9B, it reaches comparable or
higher pass@2 than our reproduction of HarnessX (71.0 against 67.3 on average)
at 13.0$\times$ lower mean API cost, from one run per benchmark. Its inductive
harnesses keep their gains on unseen test tasks, and a handcrafted seed did not
improve on the neutral start (Appendix~\ref{app:limitations} discusses
limitations and future work).

\subsection*{AI use statement}
We used generative AI tools heavily to polish the manuscript and to correct
non-native English expressions. For the appendix, AI agents also drafted text,
which we read and verified. We developed the meta-agents' instructions with the
help of AI agents, but we verified sensitive content by hand, in particular that
no instruction leaks information about a benchmark. AI agents also ran and
monitored our experiments, while we verified the reported results and the
evaluation code ourselves. We wrote the instructions for the supplementary code
with an AI agent and ran the code on our own server to confirm that it works.
Apart from these uses, \method{} itself runs on LLMs: the agent and all four
meta-agents are language models, as Sections~\ref{sec:prelim}
and~\ref{sec:method} describe. We take responsibility for the final content of
this work, including text, claims or artifacts produced with the aid of
generative AI.

\subsection*{Ethics statement}
This work evaluates agents on public benchmarks inside sandboxed environments
and involves no human subjects. Agents that search the web can reach published
copies of benchmark answers. Appendix~\ref{app:benchmarks:gaia} describes how
we block known copies on GAIA and what exposure remains.

\subsection*{Reproducibility statement}
Section~\ref{sec:method} and Appendix~\ref{app:method} specify the loop, the
meta-agents, the gate and every setting, and Appendix~\ref{app:method:neutral} gives
the neutral harness verbatim. Appendix~\ref{app:benchmarks} gives the task
sets, the grading protocol, the cost accounting and the HarnessX reproduction.
As supplementary material, we attach a minimal set of our code that runs the
\method{} experiments. We will release the full code on GitHub upon
acceptance.


\bibliography{references}

@misc{harnessx,
  title         = {{HarnessX}: A Composable, Adaptive, and Evolvable Agent Harness Foundry},
  author        = {Chen, Tingyang and Lu, Shuo and Zhao, Kang and Meng, Weicheng and Teng, Hanlin and Li, Tianhao and Li, Chao and Liu, Xule and Liang, Jian and Zhang, Zhizhong and Xie, Yuan and Qu, Heng and Shao, Kun and Luan, Jian},
  year          = {2026},
  eprint        = {2606.14249},
  archivePrefix = {arXiv},
  primaryClass  = {cs.AI},
  url           = {https://arxiv.org/abs/2606.14249}
}

@misc{evobench,
  title         = {{Evo-Bench}: Can Language Models Improve Agent Harness?},
  author        = {Huang, Lisheng and Yang, Chen and Zhou, Hao and Song, Huatong and Chen, Zongchao and Le, Ran and Song, Yang and Zhao, Wayne Xin and Zhang, Tao},
  year          = {2026},
  eprint        = {2608.09096},
  archivePrefix = {arXiv},
  primaryClass  = {cs.CL},
  url           = {https://arxiv.org/abs/2608.09096}
}

@misc{autoharness,
  title         = {{AutoHarness}: improving {LLM} agents by automatically synthesizing a code harness},
  author        = {Lou, Xinghua and L{\'a}zaro-Gredilla, Miguel and Dedieu, Antoine and Wendelken, Carter and Lehrach, Wolfgang and Murphy, Kevin P.},
  year          = {2026},
  eprint        = {2603.03329},
  archivePrefix = {arXiv},
  primaryClass  = {cs.CL},
  url           = {https://arxiv.org/abs/2603.03329},
  note          = {Presented at the ICLR 2026 Workshop on AI with Recursive Self-Improvement}
}

@inproceedings{metaharness,
  title     = {{Meta-Harness}: End-to-End Optimization of Model Harnesses},
  author    = {Lee, Yoonho and Nair, Roshen S. and Zhang, Qizheng and Lee, Kangwook and Khattab, Omar and Finn, Chelsea},
  booktitle = {Third Conference on Language Modeling},
  year      = {2026},
  url       = {https://arxiv.org/abs/2603.28052}
}

@inproceedings{gaia,
  title     = {{GAIA}: a benchmark for {General AI Assistants}},
  author    = {Mialon, Gr\'egoire and Fourrier, Cl\'ementine and Wolf, Thomas and LeCun, Yann and Scialom, Thomas},
  booktitle = {International Conference on Learning Representations (ICLR)},
  year      = {2024},
  pages     = {9025--9049},
  url       = {https://openreview.net/forum?id=fibxvahvs3}
}

@inproceedings{alfworld,
  title     = {{ALFWorld}: Aligning Text and Embodied Environments for Interactive Learning},
  author    = {Shridhar, Mohit and Yuan, Xingdi and C\^ot\'e, Marc-Alexandre and Bisk, Yonatan and Trischler, Adam and Hausknecht, Matthew},
  booktitle = {International Conference on Learning Representations (ICLR)},
  year      = {2021},
  url       = {https://openreview.net/forum?id=VpNENmFBEEg}
}

@inproceedings{webshop,
  title     = {{WebShop}: Towards Scalable Real-World Web Interaction with Grounded Language Agents},
  author    = {Yao, Shunyu and Chen, Howard and Yang, John and Narasimhan, Karthik},
  booktitle = {Advances in Neural Information Processing Systems},
  editor    = {S. Koyejo and S. Mohamed and A. Agarwal and D. Belgrave and K. Cho and A. Oh},
  volume    = {35},
  pages     = {20744--20757},
  publisher = {Curran Associates, Inc.},
  year      = {2022},
  doi       = {10.52202/068431-1508},
  url       = {https://proceedings.neurips.cc/paper_files/paper/2022/file/82ad13ec01f9fe44c01cb91814fd7b8c-Paper-Conference.pdf}
}

@inproceedings{appworld,
    title     = "{A}pp{W}orld: A Controllable World of Apps and People for Benchmarking Interactive Coding Agents",
    author    = "Trivedi, Harsh and Khot, Tushar and Hartmann, Mareike and Manku, Ruskin and Dong, Vinty and Li, Edward and Gupta, Shashank and Sabharwal, Ashish and Balasubramanian, Niranjan",
    editor    = "Ku, Lun-Wei and Martins, Andre and Srikumar, Vivek",
    booktitle = "Proceedings of the 62nd Annual Meeting of the Association for Computational Linguistics (Volume 1: Long Papers)",
    month     = aug,
    year      = "2024",
    address   = "Bangkok, Thailand",
    publisher = "Association for Computational Linguistics",
    url       = "https://aclanthology.org/2024.acl-long.850/",
    doi       = "10.18653/v1/2024.acl-long.850",
    pages     = "16022--16076"
}

@inproceedings{swebench,
  title     = {{SWE}-bench: Can Language Models Resolve Real-World {GitHub} Issues?},
  author    = {Jimenez, Carlos E. and Yang, John and Wettig, Alexander and Yao, Shunyu and Pei, Kexin and Press, Ofir and Narasimhan, Karthik},
  booktitle = {The Twelfth International Conference on Learning Representations},
  year      = {2024},
  url       = {https://openreview.net/forum?id=VTF8yNQM66}
}

@inproceedings{gigpo,
  author    = {Feng, Lang and Xue, Zhenghai and Liu, Tingcong and An, Bo},
  title     = {Group-in-Group Policy Optimization for {LLM} Agent Training},
  booktitle = {Advances in Neural Information Processing Systems},
  volume    = {38},
  pages     = {46375--46408},
  publisher = {Curran Associates, Inc.},
  year      = {2025},
  doi       = {10.52202/085713-1544},
  url       = {https://proceedings.neurips.cc/paper_files/paper/2025/hash/420c9f777c0b4f78d515e53cf74d58b2-Abstract-Conference.html}
}

@misc{chen2021codex,
      title={Evaluating Large Language Models Trained on Code}, 
      author={Mark Chen and Jerry Tworek and Heewoo Jun and Qiming Yuan and Henrique Ponde de Oliveira Pinto and Jared Kaplan and Harri Edwards and Yuri Burda and Nicholas Joseph and Greg Brockman and Alex Ray and Raul Puri and Gretchen Krueger and Michael Petrov and Heidy Khlaaf and Girish Sastry and Pamela Mishkin and Brooke Chan and Scott Gray and Nick Ryder and Mikhail Pavlov and Alethea Power and Lukasz Kaiser and Mohammad Bavarian and Clemens Winter and Philippe Tillet and Felipe Petroski Such and Dave Cummings and Matthias Plappert and Fotios Chantzis and Elizabeth Barnes and Ariel Herbert-Voss and William Hebgen Guss and Alex Nichol and Alex Paino and Nikolas Tezak and Jie Tang and Igor Babuschkin and Suchir Balaji and Shantanu Jain and William Saunders and Christopher Hesse and Andrew N. Carr and Jan Leike and Josh Achiam and Vedant Misra and Evan Morikawa and Alec Radford and Matthew Knight and Miles Brundage and Mira Murati and Katie Mayer and Peter Welinder and Bob McGrew and Dario Amodei and Sam McCandlish and Ilya Sutskever and Wojciech Zaremba},
      year={2021},
      eprint={2107.03374},
      archivePrefix={arXiv},
      primaryClass={cs.LG},
      url={https://arxiv.org/abs/2107.03374}, 
}

@misc{claudecode,
  author       = {{Anthropic}},
  title        = {{Claude Code} overview},
  howpublished = {Claude Code documentation},
  year         = {2026},
  url          = {https://code.claude.com/docs/en/overview},
  note         = {Undated documentation page. Accessed 2026-09-23}
}

@inproceedings{swerl,
  title     = {{SWE-RL}: Advancing {LLM} Reasoning via Reinforcement Learning on Open Software Evolution},
  author    = {Wei, Yuxiang and Duchenne, Olivier and Copet, Jade and Carbonneaux, Quentin and Zhang, Lingming and Fried, Daniel and Synnaeve, Gabriel and Singh, Rishabh and Wang, Sida},
  booktitle = {Advances in Neural Information Processing Systems},
  editor    = {D. Belgrave and C. Zhang and H. Lin and R. Pascanu and P. Koniusz and M. Ghassemi and N. Chen},
  volume    = {38},
  pages     = {78500--78525},
  publisher = {Curran Associates, Inc.},
  year      = {2025},
  doi       = {10.52202/085713-2629},
  url       = {https://proceedings.neurips.cc/paper_files/paper/2025/file/7107d4d2e837bde2171c6b71b5bde954-Paper-Conference.pdf}
}

@inproceedings{adas,
  title     = {Automated Design of Agentic Systems},
  author    = {Hu, Shengran and Lu, Cong and Clune, Jeff},
  booktitle = {International Conference on Learning Representations},
  editor    = {Y. Yue and A. Garg and N. Peng and F. Sha and R. Yu},
  volume    = {2025},
  pages     = {21344--21377},
  year      = {2025},
  url       = {https://proceedings.iclr.cc/paper_files/paper/2025/file/36b7acf6f6010652b3f2a433774a66fe-Paper-Conference.pdf}
}

@misc{sica,
  title         = {A Self-Improving Coding Agent},
  author        = {Maxime Robeyns and Martin Szummer and Laurence Aitchison},
  year          = {2025},
  eprint        = {2504.15228},
  archivePrefix = {arXiv},
  primaryClass  = {cs.AI},
  url           = {https://arxiv.org/abs/2504.15228},
  note          = {Presented at the ICLR 2025 Workshop on Scaling Self-Improving Foundation Models}
}

@inproceedings{dgm,
  title     = {Darwin {G\"{o}del} Machine: Open-Ended Evolution of Self-Improving Agents},
  author    = {Zhang, Jenny and Hu, Shengran and Lu, Cong and Lange, Robert and Clune, Jeff},
  booktitle = {International Conference on Learning Representations},
  editor    = {C. Vondrick and B. Hariharan and C. Raffel and L. Pinto and D. Yang and A. Faust},
  volume    = {2026},
  pages     = {104223--104294},
  year      = {2026},
  url       = {https://proceedings.iclr.cc/paper_files/paper/2026/file/aa5f5e6eb6f613ec412f1d948dfa21a5-Paper-Conference.pdf}
}

@misc{deepswe,
  title        = {{DeepSWE}: Training a Fully Open-sourced, State-of-the-Art Coding Agent by Scaling {RL}},
  author       = {Luo, Michael and Jain, Naman and Singh, Jaskirat and Tan, Sijun and Patel, Ameen and Wu, Qingyang and Ariyak, Alpay and Cai, Colin and Venkat, Tarun and Zhu, Shang and Athiwaratkun, Ben and Roongta, Manan and Zhang, Ce and Li, Li Erran and Popa, Raluca Ada and Sen, Koushik and Stoica, Ion},
  year         = {2025},
  month        = jul,
  howpublished = {Together AI Blog},
  url          = {https://www.together.ai/blog/deepswe},
  note         = {Published 2 July 2025}
}

@inproceedings{gepa,
  title     = {{GEPA}: Reflective Prompt Evolution Can Outperform Reinforcement Learning},
  author    = {Agrawal, Lakshya A. and Tan, Shangyin and Soylu, Dilara and Ziems, Noah and Khare, Rishi and Opsahl-Ong, Krista and Singhvi, Arnav and Shandilya, Herumb and Ryan, Michael J. and Jiang, Meng and Potts, Christopher and Sen, Koushik and Dimakis, Alexandros G. and Stoica, Ion and Klein, Dan and Zaharia, Matei and Khattab, Omar},
  booktitle = {International Conference on Learning Representations},
  editor    = {C. Vondrick and B. Hariharan and C. Raffel and L. Pinto and D. Yang and A. Faust},
  volume    = {2026},
  pages     = {8479--8565},
  year      = {2026},
  url       = {https://proceedings.iclr.cc/paper_files/paper/2026/file/0e9e708b6f48e14fd0ac29e167413f76-Paper-Conference.pdf}
}

@misc{rethinkharness,
  title         = {Rethinking the Evaluation of Harness Evolution for Agents},
  author        = {Yike Wang and Huaisheng Zhu and Zhengyu Hu and Yige Yuan and Zhengyu Chen and Shakti Senthil and Hannaneh Hajishirzi and Yulia Tsvetkov and Pradeep Dasigi and Teng Xiao},
  year          = {2026},
  eprint        = {2607.12227},
  archivePrefix = {arXiv},
  primaryClass  = {cs.AI},
  url           = {https://arxiv.org/abs/2607.12227},
}

@inproceedings{agenticrpo,
  title     = {Agentic Reinforced Policy Optimization},
  author    = {Dong, Guanting and Mao, Hangyu and Ma, Kai and Bao, Licheng and Chen, Yifei and Wang, Zhongyuan and Chen, Zhongxia and Du, Jiazhen and Wang, Huiyang and Zhang, Fuzheng and Zhou, Guorui and Zhu, Yutao and Wen, Ji-Rong and Dou, Zhicheng},
  booktitle = {International Conference on Learning Representations},
  editor    = {C. Vondrick and B. Hariharan and C. Raffel and L. Pinto and D. Yang and A. Faust},
  volume    = {2026},
  pages     = {16981--17017},
  year      = {2026},
  url       = {https://proceedings.iclr.cc/paper_files/paper/2026/file/1c58e53bdf1fb045440256fd567531ae-Paper-Conference.pdf}
}

@inproceedings{agentgymrl,
  title     = {{AgentGym-RL}: An Open-Source Framework to Train {LLM} Agents for Long-Horizon Decision Making via Multi-Turn {RL}},
  author    = {Xi, Zhiheng and Huang, Jixuan and Liao, Chenyang and Huang, Baodai and Liu, Jiaqi and Guo, Honglin and Yang, Yajie and Zheng, Rui and Ye, Junjie and Zhang, Jiazheng and Chen, Wenxiang and He, Wei and Ding, Yiwen and Li, Guanyu and Chen, Zehui and Du, Zhengyin and Yao, Xuesong and Xu, Yufei and Chen, Jiecao and Gui, Tao and Wu, Zuxuan and Zhang, Qi and Huang, Xuanjing and Jiang, Yu-Gang},
  booktitle = {International Conference on Learning Representations},
  editor    = {C. Vondrick and B. Hariharan and C. Raffel and L. Pinto and D. Yang and A. Faust},
  volume    = {2026},
  pages     = {12685--12727},
  year      = {2026},
  url       = {https://proceedings.iclr.cc/paper_files/paper/2026/file/1571ce1ff735be4551db50887043726e-Paper-Conference.pdf}
}

@inproceedings{webagentr1,
  title     = {{W}eb{A}gent-{R}1: Training Web Agents via End-to-End Multi-Turn Reinforcement Learning},
  author    = {Wei, Zhepei and Yao, Wenlin and Liu, Yao and Zhang, Weizhi and Lu, Qin and Qiu, Liang and Yu, Changlong and Xu, Puyang and Zhang, Chao and Yin, Bing and Yun, Hyokun and Li, Lihong},
  editor    = {Christodoulopoulos, Christos and Chakraborty, Tanmoy and Rose, Carolyn and Peng, Violet},
  booktitle = {Proceedings of the 2025 Conference on Empirical Methods in Natural Language Processing},
  month     = nov,
  year      = {2025},
  address   = {Suzhou, China},
  publisher = {Association for Computational Linguistics},
  url       = {https://aclanthology.org/2025.emnlp-main.401/},
  doi       = {10.18653/v1/2025.emnlp-main.401},
  pages     = {7909--7928},
  isbn      = {979-8-89176-332-6}
}

@article{agenticrlsurvey,
  title   = {The Landscape of Agentic Reinforcement Learning for {LLM}s: A Survey},
  author  = {Zhang, Guibin and Geng, Hejia and Yu, Xiaohang and Yin, Zhenfei and Zhang, Zaibin and Tan, Zelin and Zhou, Heng and Li, Zhongzhi and Xue, Xiangyuan and Li, Yijiang and Zhou, Yifan and Chen, Yang and Zhang, Chen and Fan, Yutao and Wang, Zihu and Huang, Songtao and Piedrahita-Velez, Francisco and Liao, Yue and Wang, Hongru and Yang, Mengyue and Ji, Heng and Wang, Jun and Yan, Shuicheng and Torr, Philip and Bai, Lei},
  journal = {Transactions on Machine Learning Research},
  issn    = {2835-8856},
  year    = {2026},
  url     = {https://openreview.net/forum?id=RY19y2RI1O}
}

@inproceedings{agenttuning,
    title = "{A}gent{T}uning: Enabling Generalized Agent Abilities for {LLM}s",
    author = "Zeng, Aohan  and
      Liu, Mingdao  and
      Lu, Rui  and
      Wang, Bowen  and
      Liu, Xiao  and
      Dong, Yuxiao  and
      Tang, Jie",
    editor = "Ku, Lun-Wei  and
      Martins, Andre  and
      Srikumar, Vivek",
    booktitle = "Findings of the Association for Computational Linguistics: ACL 2024",
    month = aug,
    year = "2024",
    address = "Bangkok, Thailand",
    publisher = "Association for Computational Linguistics",
    url = "https://aclanthology.org/2024.findings-acl.181/",
    doi = "10.18653/v1/2024.findings-acl.181",
    pages = "3053--3077"
}

@misc{fireact,
  title         = {{FireAct}: Toward Language Agent Fine-tuning},
  author        = {Chen, Baian and Shu, Chang and Shareghi, Ehsan and Collier, Nigel and Narasimhan, Karthik and Yao, Shunyu},
  year          = {2023},
  eprint        = {2310.05915},
  archivePrefix = {arXiv},
  primaryClass  = {cs.CL},
  url           = {https://arxiv.org/abs/2310.05915}
}

@inproceedings{agentbank,
    title = "{A}gent{B}ank: Towards Generalized {LLM} Agents via Fine-Tuning on 50000+ Interaction Trajectories",
    author = "Song, Yifan  and
      Xiong, Weimin  and
      Zhao, Xiutian  and
      Zhu, Dawei  and
      Wu, Wenhao  and
      Wang, Ke  and
      Li, Cheng  and
      Peng, Wei  and
      Li, Sujian",
    editor = "Al-Onaizan, Yaser  and
      Bansal, Mohit  and
      Chen, Yun-Nung",
    booktitle = "Findings of the Association for Computational Linguistics: EMNLP 2024",
    month = nov,
    year = "2024",
    address = "Miami, Florida, USA",
    publisher = "Association for Computational Linguistics",
    url = "https://aclanthology.org/2024.findings-emnlp.116/",
    doi = "10.18653/v1/2024.findings-emnlp.116",
    pages = "2124--2141"
}

@inproceedings{agentflan,
    title = "Agent-{FLAN}: Designing Data and Methods of Effective Agent Tuning for Large Language Models",
    author = "Chen, Zehui  and
      Liu, Kuikun  and
      Wang, Qiuchen  and
      Zhang, Wenwei  and
      Liu, Jiangning  and
      Lin, Dahua  and
      Chen, Kai  and
      Zhao, Feng",
    editor = "Ku, Lun-Wei  and
      Martins, Andre  and
      Srikumar, Vivek",
    booktitle = "Findings of the Association for Computational Linguistics: ACL 2024",
    month = aug,
    year = "2024",
    address = "Bangkok, Thailand",
    publisher = "Association for Computational Linguistics",
    url = "https://aclanthology.org/2024.findings-acl.557/",
    doi = "10.18653/v1/2024.findings-acl.557",
    pages = "9354--9366"
}

@inproceedings{autogen,
  title     = {{AutoGen}: Enabling Next-Gen {LLM} Applications via Multi-Agent Conversations},
  author    = {Wu, Qingyun and Bansal, Gagan and Zhang, Jieyu and Wu, Yiran and Li, Beibin and Zhu, Erkang and Jiang, Li and Zhang, Xiaoyun and Zhang, Shaokun and Liu, Jiale and Awadallah, Ahmed Hassan and White, Ryen W. and Burger, Doug and Wang, Chi},
  booktitle = {First Conference on Language Modeling},
  year      = {2024},
  url       = {https://openreview.net/forum?id=BAakY1hNKS}
}

@inproceedings{metagpt,
  title     = {{MetaGPT}: Meta Programming for A Multi-Agent Collaborative Framework},
  author    = {Hong, Sirui and Zhuge, Mingchen and Chen, Jonathan and Zheng, Xiawu and Cheng, Yuheng and Wang, Jinlin and Zhang, Ceyao and Wang, Zili and Yau, Steven and Lin, Zijuan and Zhou, Liyang and Ran, Chenyu and Xiao, Lingfeng and Wu, Chenglin and Schmidhuber, J{\"u}rgen},
  booktitle = {International Conference on Learning Representations},
  editor    = {B. Kim and Y. Yue and S. Chaudhuri and K. Fragkiadaki and M. Khan and Y. Sun},
  pages     = {23247--23275},
  volume    = {2024},
  year      = {2024},
  url       = {https://openreview.net/forum?id=VtmBAGCN7o}
}

@misc{magenticone,
  title         = {{Magentic-One}: A Generalist Multi-Agent System for Solving Complex Tasks},
  author        = {Fourney, Adam and Bansal, Gagan and Mozannar, Hussein and Tan, Cheng and Salinas, Eduardo and Zhu, Erkang and Niedtner, Friederike and Proebsting, Grace and Bassman, Griffin and Gerrits, Jack and Alber, Jacob and Chang, Peter and Loynd, Ricky and West, Robert and Dibia, Victor and Awadallah, Ahmed and Kamar, Ece and Hosn, Rafah and Amershi, Saleema},
  year          = {2024},
  eprint        = {2411.04468},
  archivePrefix = {arXiv},
  primaryClass  = {cs.AI},
  url           = {https://arxiv.org/abs/2411.04468}
}

@inproceedings{conductor,
  title     = {Learning to Orchestrate Agents in Natural Language with the Conductor},
  author    = {Nielsen, Stefan and Cetin, Edoardo and Schwendeman, Peter and Sun, Qi and Xu, Jinglue and Tang, Yujin},
  booktitle = {International Conference on Learning Representations},
  editor    = {C. Vondrick and B. Hariharan and C. Raffel and L. Pinto and D. Yang and A. Faust},
  pages     = {135686--135724},
  volume    = {2026},
  year      = {2026},
  url       = {https://openreview.net/forum?id=U23A2BUKYt}
}

@inproceedings{sweagent,
  title     = {{SWE}-agent: Agent-Computer Interfaces Enable Automated Software Engineering},
  author    = {Yang, John and Jimenez, Carlos E. and Wettig, Alexander and Lieret, Kilian and Yao, Shunyu and Narasimhan, Karthik and Press, Ofir},
  booktitle = {Advances in Neural Information Processing Systems},
  editor    = {A. Globerson and L. Mackey and D. Belgrave and A. Fan and U. Paquet and J. Tomczak and C. Zhang},
  volume    = {37},
  pages     = {50528--50652},
  publisher = {Curran Associates, Inc.},
  year      = {2024},
  doi       = {10.52202/079017-1601},
  url       = {https://proceedings.neurips.cc/paper_files/paper/2024/hash/5a7c947568c1b1328ccc5230172e1e7c-Abstract-Conference.html}
}

@inproceedings{openhands,
  title     = {{OpenHands}: An Open Platform for {AI} Software Developers as Generalist Agents},
  author    = {Wang, Xingyao and Li, Boxuan and Song, Yufan and Xu, Frank F. and Tang, Xiangru and Zhuge, Mingchen and Pan, Jiayi and Song, Yueqi and Li, Bowen and Singh, Jaskirat and Tran, Hoang H. and Li, Fuqiang and Ma, Ren and Zheng, Mingzhang and Qian, Bill and Shao, Yanjun and Muennighoff, Niklas and Zhang, Yizhe and Hui, Binyuan and Lin, Junyang and Brennan, Robert and Peng, Hao and Ji, Heng and Neubig, Graham},
  booktitle = {International Conference on Learning Representations},
  editor    = {Y. Yue and A. Garg and N. Peng and F. Sha and R. Yu},
  volume    = {2025},
  pages     = {65882--65919},
  year      = {2025},
  url       = {https://proceedings.iclr.cc/paper_files/paper/2025/file/a4b6ad6b48850c0c331d1259fc66a69c-Paper-Conference.pdf}
}

@inproceedings{terminalbench,
  title     = {{Terminal-Bench}: Benchmarking Agents on Hard, Realistic Tasks in Command Line Interfaces},
  author    = {Merrill, Mike A. and Shaw, Alexander G. and Carlini, Nicholas and Li, Boxuan and Raj, Harsh and Bercovich, Ivan and Shi, Lin and Shin, Jeong Yeon and Walshe, Thomas and Buchanan, E. Kelly and Shen, Junhong and Ye, Guanghao and Lin, Haowei and Poulos, Jason and Wang, Maoyu and Nezhurina, Marianna and Lu, Di and Menis Mastromichalakis, Orfeas and Xu, Zhiwei and Chen, Zizhao and Liu, Yue and Zhang, Robert and Chen, Leon Liangyu and Kashyap, Anurag and Uslu, Jan-Lucas and Li, Jeffrey and Wu, Jianbo and Yan, Minghao and Bian, Song and Sharma, Vedang and Sun, Ke and Dillmann, Steven and Anand, Akshay and Lanpouthakoun, Andrew and Koopah, Bardia and Hu, Changran and Guha, Etash and Dreiman, Gabriel H. S. and Zhu, Jiacheng and Krauth, Karl and Zhong, Li and Muennighoff, Niklas and Amanfu, Robert and Tan, Shangyin and Pimpalgaonkar, Shreyas and Aggarwal, Tushar and Lin, Xiangning and Lan, Xin and Zhao, Xuandong and Liang, Yiqing and Wang, Yuanli and Wang, Zilong and Zhou, Changzhi and Heineman, David and Liu, Hange and Trivedi, Harsh and Yang, John and Lin, Junhong and Shetty, Manish and Yang, Michael and Omi, Nabil and Raoof, Negin and Li, Shanda and Zhuo, Terry Yue and Lin, Wuwei and Dai, Yiwei and Wang, Yuxin and Chai, Wenhao and Zhou, Shang and Wahdany, Dariush and She, Ziyu and Hu, Jiaming and Dong, Zhikang and Zhu, Yuxuan and Cui, Sasha and Saiyed, Ahson and Kolbeinsson, Arinbj{\"o}rn and Hu, Jesse and Rytting, Christopher Michael and Marten, Ryan and Wang, Yixin and Jitsev, Jenia and Dimakis, Alex and Konwinski, Andy and Schmidt, Ludwig},
  booktitle = {International Conference on Learning Representations},
  editor    = {C. Vondrick and B. Hariharan and C. Raffel and L. Pinto and D. Yang and A. Faust},
  volume    = {2026},
  pages     = {40903--40986},
  year      = {2026},
  url       = {https://proceedings.iclr.cc/paper_files/paper/2026/file/444a3737adaee10d86ad2ef5f74468e6-Paper-Conference.pdf}
}

@misc{claudecodeharness,
  author       = {{Anthropic}},
  title        = {How {Claude Code} works},
  howpublished = {Claude Code documentation},
  year         = {2026},
  url          = {https://code.claude.com/docs/en/how-claude-code-works},
  note         = {Undated documentation page. Accessed 2026-09-23}
}

@inproceedings{dspy,
  title         = {{DSPy}: Compiling Declarative Language Model Calls into State-of-the-Art Pipelines},
  author        = {Omar Khattab and Arnav Singhvi and Paridhi Maheshwari and Zhiyuan Zhang and Keshav Santhanam and Sri Vardhamanan and Saiful Haq and Ashutosh Sharma and Thomas T. Joshi and Hanna Moazam and Heather Miller and Matei Zaharia and Christopher Potts},
  booktitle     = {The Twelfth International Conference on Learning Representations},
  year          = {2024},
  note          = {Spotlight},
  url           = {https://openreview.net/forum?id=sY5N0zY5Od},
  eprint        = {2310.03714},
  archivePrefix = {arXiv}
}

@inproceedings{mipro,
    title = "Optimizing Instructions and Demonstrations for Multi-Stage Language Model Programs",
    author = "Opsahl-Ong, Krista  and
      Ryan, Michael J  and
      Purtell, Josh  and
      Broman, David  and
      Potts, Christopher  and
      Zaharia, Matei  and
      Khattab, Omar",
    editor = "Al-Onaizan, Yaser  and
      Bansal, Mohit  and
      Chen, Yun-Nung",
    booktitle = "Proceedings of the 2024 Conference on Empirical Methods in Natural Language Processing",
    month = nov,
    year = "2024",
    address = "Miami, Florida, USA",
    publisher = "Association for Computational Linguistics",
    url = "https://aclanthology.org/2024.emnlp-main.525/",
    doi = "10.18653/v1/2024.emnlp-main.525",
    pages = "9340--9366"
}

@inproceedings{opro,
  title         = {Large Language Models as Optimizers},
  author        = {Chengrun Yang and Xuezhi Wang and Yifeng Lu and Hanxiao Liu and Quoc V. Le and Denny Zhou and Xinyun Chen},
  booktitle     = {The Twelfth International Conference on Learning Representations},
  year          = {2024},
  url           = {https://openreview.net/forum?id=Bb4VGOWELI},
  eprint        = {2309.03409},
  archivePrefix = {arXiv}
}

@misc{textgrad,
  title         = {TextGrad: Automatic "Differentiation" via Text},
  author        = {Mert Yuksekgonul and Federico Bianchi and Joseph Boen and Sheng Liu and Zhi Huang and Carlos Guestrin and James Zou},
  year          = {2024},
  eprint        = {2406.07496},
  archivePrefix = {arXiv},
  primaryClass  = {cs.CL},
  url           = {https://arxiv.org/abs/2406.07496},
  note          = {A journal version appeared under a different title; not verified here}
}

@inproceedings{ace,
  title     = {Agentic Context Engineering: Evolving Contexts for Self-Improving Language Models},
  author    = {Zhang, Qizheng and Hu, Changran and Upasani, Shubhangi and Ma, Boyuan and Hong, Fenglu and Kamanuru, Vamsidhar and Rainton, Jay and Wu, Chen and Ji, Mengmeng and Li, Hanchen and Thakker, Urmish and Zou, James and Olukotun, Kunle},
  booktitle = {International Conference on Learning Representations},
  editor    = {C. Vondrick and B. Hariharan and C. Raffel and L. Pinto and D. Yang and A. Faust},
  volume    = {2026},
  pages     = {86069--86100},
  year      = {2026},
  url       = {https://proceedings.iclr.cc/paper_files/paper/2026/file/8a94ff6f922d995d7d3f4ebf4143e442-Paper-Conference.pdf}
}

@inproceedings{reflexion,
  title         = {Reflexion: Language Agents with Verbal Reinforcement Learning},
  author        = {Noah Shinn and Federico Cassano and Edward Berman and Ashwin Gopinath and Karthik Narasimhan and Shunyu Yao},
  booktitle     = {Advances in Neural Information Processing Systems 36},
  year          = {2023},
  url           = {https://openreview.net/forum?id=vAElhFcKW6},
  eprint        = {2303.11366},
  archivePrefix = {arXiv}
}

@inproceedings{expel,
  title     = {{ExpeL}: {LLM} Agents Are Experiential Learners},
  author    = {Andrew Zhao and Daniel Huang and Quentin Xu and Matthieu Lin and Yong-Jin Liu and Gao Huang},
  booktitle = {Proceedings of the AAAI Conference on Artificial Intelligence},
  volume    = {38},
  pages     = {19632--19642},
  year      = {2024},
  doi       = {10.1609/aaai.v38i17.29936},
  url       = {https://ojs.aaai.org/index.php/AAAI/article/view/29936}
}

@inproceedings{automanual,
  title         = {AutoManual: Constructing Instruction Manuals by LLM Agents via Interactive Environmental Learning},
  author        = {Minghao Chen and Yihang Li and Yanting Yang and Shiyu Yu and Binbin Lin and Xiaofei He},
  booktitle     = {Advances in Neural Information Processing Systems 37},
  year          = {2024},
  url           = {https://openreview.net/forum?id=Pwl9n4zlf5},
  eprint        = {2405.16247},
  archivePrefix = {arXiv}
}

@inproceedings{awm,
  title         = {Agent Workflow Memory},
  author        = {Zora Zhiruo Wang and Jiayuan Mao and Daniel Fried and Graham Neubig},
  booktitle     = {Proceedings of the 42nd International Conference on Machine Learning},
  year          = {2025},
  note          = {Venue per OpenReview/dblp record (ICML 2025)},
  url           = {https://arxiv.org/abs/2409.07429},
  eprint        = {2409.07429},
  archivePrefix = {arXiv}
}

@inproceedings{dynamiccheatsheet,
  title         = {Dynamic Cheatsheet: Test-Time Learning with Adaptive Memory},
  author        = {Suzgun, Mirac and Yuksekgonul, Mert and Bianchi, Federico and Jurafsky, Dan and Zou, James},
  booktitle     = {Proceedings of the 19th Conference of the {E}uropean Chapter of the {A}ssociation for {C}omputational {L}inguistics (Volume 1: Long Papers)},
  pages         = {7080--7106},
  year          = {2026},
  month         = {mar},
  publisher     = {Association for Computational Linguistics},
  address       = {Rabat, Morocco},
  doi           = {10.18653/v1/2026.eacl-long.333},
  eprint        = {2504.07952},
  archivePrefix = {arXiv},
  url           = {https://aclanthology.org/2026.eacl-long.333/}
}

@inproceedings{reasoningbank,
  title         = {ReasoningBank: Scaling Agent Self-Evolving with Reasoning Memory},
  author        = {Siru Ouyang and Jun Yan and I-Hung Hsu and Yanfei Chen and Ke Jiang and Zifeng Wang and Rujun Han and Long T. Le and Samira Daruki and Xiangru Tang and Vishy Tirumalashetty and George Lee and Mahsan Rofouei and Hangfei Lin and Jiawei Han and Chen-Yu Lee and Tomas Pfister},
  booktitle     = {The Fourteenth International Conference on Learning Representations},
  year          = {2026},
  url           = {https://openreview.net/forum?id=jL7fwchScm},
  eprint        = {2509.25140},
  archivePrefix = {arXiv}
}

@misc{harnesscompass,
  title         = {{HarnessCompass}: Guiding Automatic Harness Evolution toward Generalizable and Effective Agent Harnesses},
  author        = {Zhang, Luan and Zhou, Ruochen and Song, Dandan and Chen, Zhengyu and Tian, Yuhang and Yang, Jun and Ma, Huipeng and Li, Chenhao and Feng, Guangyuan and Li, Xudong and Jin, Yizhou and Xu, Yan},
  year          = {2026},
  eprint        = {2608.01918},
  archivePrefix = {arXiv},
  primaryClass  = {cs.LG},
  url           = {https://arxiv.org/abs/2608.01918}
}

@misc{selfharness,
  title         = {{Self-Harness}: Harnesses That Improve Themselves},
  author        = {Zhang, Hangfan and Zhang, Shao and Li, Kangcong and Zhang, Chen and Chen, Yang and Zhang, Yiqun and Bai, Lei and Hu, Shuyue},
  year          = {2026},
  eprint        = {2606.09498},
  archivePrefix = {arXiv},
  primaryClass  = {cs.CL},
  url           = {https://arxiv.org/abs/2606.09498}
}

@misc{betterharness,
  title         = {Better Harnesses, Smaller Models: Building 90\% Cheaper Agents via Automated Harness Adaptation},
  author        = {Chenyang Yang and Xinran Zhao and Tongshuang Wu and Christian K{\"a}stner},
  year          = {2026},
  eprint        = {2607.08938},
  archivePrefix = {arXiv},
  primaryClass  = {cs.SE},
  url           = {https://arxiv.org/abs/2607.08938}
}

@misc{ahe,
  title         = {Agentic Harness Engineering: Observability-Driven Automatic Evolution of Coding-Agent Harnesses},
  author        = {Lin, Jiahang and Liu, Shichun and Pan, Chengjun and Lin, Lizhi and Dou, Shihan and Xi, Zhiheng and Huang, Xuanjing and Yan, Hang and Han, Zhenhua and Gui, Tao and Jiang, Yu-Gang},
  year          = {2026},
  eprint        = {2604.25850},
  archivePrefix = {arXiv},
  primaryClass  = {cs.CL},
  url           = {https://arxiv.org/abs/2604.25850}
}

@misc{harnessupdating,
  title         = {Harness Updating Is Not Harness Benefit: Disentangling Evolution Capabilities in Self-Evolving LLM Agents},
  author        = {Minhua Lin and Juncheng Wu and Zijun Wang and Zhan Shi and Yisi Sang and Bing He and Zewen Liu and Tianxin Wei and Zongyu Wu and Zhiwei Zhang and Dakuo Wang and Xiang Zhang and Benoit Dumoulin and Cihang Xie and Yuyin Zhou and Suhang Wang and Hanqing Lu},
  year          = {2026},
  eprint        = {2605.30621},
  archivePrefix = {arXiv},
  primaryClass  = {cs.AI},
  url           = {https://arxiv.org/abs/2605.30621}
}

@inproceedings{aflow,
  title     = {{AFlow}: Automating Agentic Workflow Generation},
  author    = {Zhang, Jiayi and Xiang, Jinyu and Yu, Zhaoyang and Teng, Fengwei and Chen, Xiong-Hui and Chen, Jiaqi and Zhuge, Mingchen and Cheng, Xin and Hong, Sirui and Wang, Jinlin and Zheng, Bingnan and Liu, Bang and Luo, Yuyu and Wu, Chenglin},
  booktitle = {International Conference on Learning Representations (ICLR)},
  editor    = {Y. Yue and A. Garg and N. Peng and F. Sha and R. Yu},
  volume    = {2025},
  pages     = {34040--34077},
  year      = {2025},
  url       = {https://openreview.net/forum?id=z5uVAKwmjf},
  note      = {Oral. arXiv:2410.10762}
}

@inproceedings{agentsquare,
  title     = {{AgentSquare}: Automatic {LLM} Agent Search in Modular Design Space},
  author    = {Shang, Yu and Li, Yu and Zhao, Keyu and Ma, Likai and Liu, Jiahe and Xu, Fengli and Li, Yong},
  booktitle = {International Conference on Learning Representations (ICLR)},
  editor    = {Y. Yue and A. Garg and N. Peng and F. Sha and R. Yu},
  volume    = {2025},
  pages     = {3841--3865},
  year      = {2025},
  url       = {https://openreview.net/forum?id=mPdmDYIQ7f},
  note      = {arXiv:2410.06153}
}

@inproceedings{maas,
  title         = {Multi-agent Architecture Search via Agentic Supernet},
  author        = {Guibin Zhang and Luyang Niu and Junfeng Fang and Kun Wang and Lei Bai and Xiang Wang},
  booktitle     = {Proceedings of the 42nd International Conference on Machine Learning},
  year          = {2025},
  note          = {Oral},
  url           = {https://openreview.net/forum?id=imcyVlzpXh},
  eprint        = {2502.04180},
  archivePrefix = {arXiv}
}

@misc{grpo,
  title         = {{DeepSeekMath}: Pushing the Limits of Mathematical Reasoning in Open Language Models},
  author        = {Zhihong Shao and Peiyi Wang and Qihao Zhu and Runxin Xu and Junxiao Song and Xiao Bi and Haowei Zhang and Mingchuan Zhang and Y. K. Li and Y. Wu and Daya Guo},
  year          = {2024},
  eprint        = {2402.03300},
  archivePrefix = {arXiv},
  primaryClass  = {cs.CL},
  url           = {https://arxiv.org/abs/2402.03300}
}

@misc{dapo,
  title         = {{DAPO}: An Open-Source {LLM} Reinforcement Learning System at Scale},
  author        = {Qiying Yu and Zheng Zhang and Ruofei Zhu and Yufeng Yuan and Xiaochen Zuo and Yu Yue and Weinan Dai and Tiantian Fan and Gaohong Liu and Lingjun Liu and Xin Liu and Haibin Lin and Zhiqi Lin and Bole Ma and Guangming Sheng and Yuxuan Tong and Chi Zhang and Mofan Zhang and Wang Zhang and Hang Zhu and Jinhua Zhu and Jiaze Chen and Jiangjie Chen and Chengyi Wang and Hongli Yu and Yuxuan Song and Xiangpeng Wei and Hao Zhou and Jingjing Liu and Wei-Ying Ma and Ya-Qin Zhang and Lin Yan and Mu Qiao and Yonghui Wu and Mingxuan Wang},
  year          = {2025},
  eprint        = {2503.14476},
  archivePrefix = {arXiv},
  primaryClass  = {cs.LG},
  url           = {https://arxiv.org/abs/2503.14476}
}

@misc{tfgrpo,
  title         = {Training-Free Group Relative Policy Optimization},
  author        = {Yuzheng Cai and Siqi Cai and Yuchen Shi and Zihan Xu and Lichao Chen and Yulei Qin and Xiaoyu Tan and Gang Li and Zongyi Li and Haojia Lin and Yong Mao and Ke Li and Xing Sun},
  year          = {2025},
  eprint        = {2510.08191},
  archivePrefix = {arXiv},
  primaryClass  = {cs.CL},
  url           = {https://arxiv.org/abs/2510.08191}
}
\bibliographystyle{iclr2027_conference}

\appendix

\clearpage
\section{Related work}
\label{app:related}
HarnessX~\citep{harnessx} is the closest work: its Opus~4.6 meta-agent evolves
harnesses on four of our five benchmarks, starting each from a handcrafted
harness, and the paper reports peak-round pass@2 on the tasks it evolves on.
Other harness optimizers likewise build knowledge of the benchmark into the
start or into the optimizer's instructions, through per-domain seeds and
skills~\citep{metaharness}, the benchmark's official agent~\citep{ace}, the
game's description~\citep{autoharness}, an exploration agent aimed at the
benchmark~\citep{ahe}, or software-repair hints~\citep{harnesscompass}.
Methods that evolve whole agents report high costs for a whole run: about
\$300 to \$500 for ADAS, which scores held-out tasks, and about \$7{,}000 and
\$22{,}000 for the self-improving coding agents SICA and
DGM~\citep{adas,sica,dgm}. Cheaper methods that score held-out tasks, such as
GEPA and Better Harnesses, are evaluated on none of our
benchmarks~\citep{gepa,betterharness}. Once search and test tasks are
separated, harness evolution in the style of AHE gains only 0.6~pp on
Terminal-Bench~2.1~\citep{rethinkharness}. \method{} targets all of these
limits at once: it starts all five benchmarks from one neutral harness, never
names the benchmark to its meta-agents, costs under \$30 per run in API calls,
and reports held-out as well as transductive scores.

Table~\ref{tab:related} compares \method{} with the closest methods on four
axes: the starting harness, what the optimizer is told about the benchmark, how
far from the evolution tasks the result is measured, and the optimizer cost
that each paper reports.

\begin{table}[H]
  \centering
  \caption{How \method{} and the closest methods start, what their optimizer is told about the benchmark, how far from the evolution tasks they are evaluated, and what optimizer cost they report. Levels: L0, evolved and graded on the same tasks; L1, unseen tasks of the same benchmark; L1.5, tasks that need unseen apps; L2, a sibling benchmark; L3, a foreign benchmark. Evo-Bench is a benchmark of evolver models rather than a method.}
  \label{tab:related}
  \footnotesize
  \setlength{\tabcolsep}{3.5pt}
  \renewcommand{\arraystretch}{1.05}
  \begin{tabular}{@{}>{\raggedright\arraybackslash}p{0.155\linewidth}>{\raggedright\arraybackslash}p{0.19\linewidth}>{\raggedright\arraybackslash}p{0.21\linewidth}>{\raggedright\arraybackslash}p{0.24\linewidth}>{\raggedright\arraybackslash}p{0.12\linewidth}@{}}
    \toprule
    Method & Starting harness & What the optimizer is told about the benchmark & Evaluation & Cost reported \\
    \midrule
    HarnessX & handcrafted for each benchmark & its benchmark-specific seeds & L0 & token budget \\
    Meta-Harness & per-domain seeds & a skill written by hand for each domain & L0 on its agentic benchmark; held-out and new datasets for classification & iterations, hours \\
    AutoHarness & one template & the game's name and description & new random instances of each game & iterations \\
    AHE & a single shell tool & an explore agent aimed at the benchmark & L0, plus one other benchmark without re-evolution & hours \\
    HarnessCompass & a single shell tool & software-repair hints in the meta prompt & L1 on one benchmark, plus a second model & iterations \\
    ACE & the benchmark's official agent & a reflector named after the benchmark & L1 offline; online on the test stream & latency, rollouts \\
    Evo-Bench & one minimal seed & the domains, in the evolver prompt & L1 & USD per run \\
    \midrule
    \method{} & one neutral harness for all five benchmarks & no benchmark identity; a filter blocks benchmark names & L0 on five benchmarks, L1 on three, L1.5 on AppWorld; L2 and L3 tried; no transfer shown & tokens and USD, also for HarnessX \\
    \bottomrule
  \end{tabular}
\end{table}


\paragraph{Training the Agent's Weights.}
Reinforcement learning trains an agent's weights on its own multi-turn
interaction~\citep{gigpo,agenticrpo,agentgymrl,webagentr1}; \citet{agenticrlsurvey}
survey the area. Such training adapts the model itself, but one run occupies a GPU cluster: SWE-RL trains on 512 GPUs for about 32 hours~\citep{swerl}, and DeepSWE on 64 GPUs for six days~\citep{deepswe}. Supervised fine-tuning on agent
trajectories~\citep{agenttuning,fireact,agentbank,agentflan} is cheaper per run
but needs trajectories from a stronger model, and it applies only to open
weights. Since \method{} changes no weight, it applies to closed agents such as Claude Code, and its evolved harnesses could be combined with either kind of
training.

\paragraph{Multi-Agent Orchestration.}
Multi-agent systems split a task among specialist agents under an orchestrator
or a fixed workflow~\citep{autogen,metagpt,magenticone,conductor}. \method{}
uses several LLM roles as well, but only offline, to write the harness of a
single agent that then runs alone at test time.

\paragraph{How Much the Harness Matters.}
With the model fixed, scores vary widely with the harness: SWE-agent's
interface solves 10.7~pp more SWE-bench instances than a plain shell with
the same model~\citep{sweagent}.
A coding agent that the DGM evolved lifts Claude~3.7 Sonnet from 19.0\% to
59.5\% on 200 SWE-bench tasks~\citep{dgm}, and one model scores from
13.9\% to 37.6\% on Terminal-Bench~2 depending on its harness~\citep{metaharness}.
Coding agents such as Claude Code are themselves harnesses around a
model~\citep{claudecodeharness,openhands}. The choice of model can still matter more than the harness~\citep{terminalbench}, which is one reason we also
evolve harnesses for a frontier agent (Table~\ref{tab:claude_code_evolite}).

\paragraph{Prompt, Context and Memory Optimization.}
Prompt optimizers tune the instructions and demonstrations of a fixed program
from feedback~\citep{dspy,mipro,opro,textgrad}. GEPA reflects on traces to
evolve prompts and beats reinforcement learning with up to 35$\times$ fewer
rollouts~\citep{gepa}, and ACE grows a playbook of strategies in an agent's
context~\citep{ace}. A second line keeps experience in memory: Reflexion
retries a task with verbal self-reflection~\citep{reflexion}, ExpeL and
AutoManual distill insights or manuals from training
tasks~\citep{expel,automanual}, and Agent Workflow Memory, Dynamic Cheatsheet
and ReasoningBank accumulate workflows, notes or strategies as a test stream
unfolds~\citep{awm,dynamiccheatsheet,reasoningbank}. \method{}'s library stores the same kinds of knowledge in typed components, next to processors and runner
settings, and a composer decides which of them the agent sees. Unlike test-stream memories, \method{} fits its library offline and freezes its harness at test time.

\paragraph{Automated Design of Agents and Harnesses.}
Earlier work searches over whole agent designs: ADAS programs new agents in
code~\citep{adas}, AFlow searches over code-represented
workflows~\citep{aflow}, AgentSquare searches a modular design
space~\citep{agentsquare}, and MaAS samples multi-agent systems from a learned
supernet~\citep{maas}. DGM and SICA let a coding agent rewrite its own code, at
about \$22{,}000 and \$7{,}000 per run~\citep{dgm,sica}. Harness evolution instead
keeps the model fixed and changes the layer around it. HarnessX, Meta-Harness,
AutoHarness, AHE, HarnessCompass and Self-Harness evolve prompts, tools and
code around a task
agent~\citep{harnessx,metaharness,autoharness,ahe,harnesscompass,selfharness},
several of them with agentic meta-agents that use file or web tools. Better
Harnesses fits harnesses for small models at \$20 per run, rollouts
included~\citep{betterharness}. Studies of the setting find that gains shrink
when search and test are separated~\citep{rethinkharness} and that a small
evolver can match a frontier one on in-place skill
edits~\citep{harnessupdating}, while Evo-Bench evaluates evolvers on held-out tasks~\citep{evobench}. \method{}'s meta-agents, by contrast, are tool-free, read a
failure-oriented sample of trajectories instead of full traces, and never see
the benchmark's name.

\clearpage
\section{Method details}
\label{app:method}

This appendix gives what Section~\ref{sec:method} and
Algorithm~\ref{alg:evolve} leave out.

\subsection{The neutral harness}
\label{app:method:neutral}
\begin{figure}[H]
  \centering
  \includegraphics[width=\textwidth]{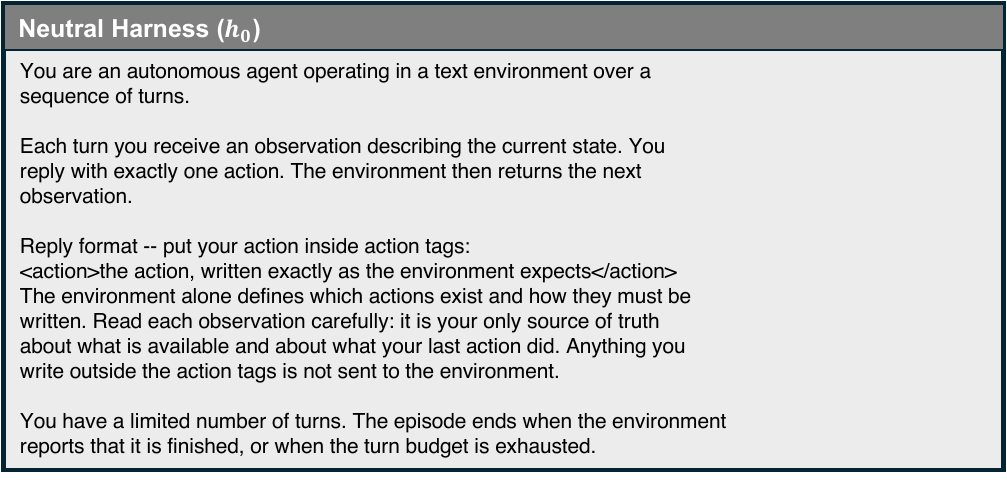}
  \caption{The neutral harness $h_0$, the system prompt with which every run starts. Line breaks inside paragraphs are added for display.}
  \label{fig:neutral}
\end{figure}

Every run starts from the system prompt in Figure~\ref{fig:neutral}, which is
788 characters long and, with the runner's reply-format block, labels and
reminder (Appendix~\ref{app:method:priors}), is the only text we write for the
agent.

On AppWorld, GAIA and SWE-bench Verified, whose actions are code, the text
changes in three places and grows to 977 characters: ``one action'' becomes
``one action block'', the example spans two lines, and one sentence says that
the text inside the tags may span several lines and runs as one unit. The
neutral harness has no processors, and its knobs take the defaults of
Table~\ref{tab:knobs}.

\subsection{The runner}
\label{app:method:runner}
The runner is a multi-turn chat loop, the same for all benchmarks and
harnesses. Its system message is $s$, and each user message holds the latest
observation, preceded by the number of turns left and followed by the list of
available actions when the knobs ask for them. The model thinks before it
replies, but its reasoning is not kept in the history. The system message and
the first observation are never dropped; past a prompt budget of 32{,}768
tokens, the oldest reply--observation pairs are dropped, while the four most
recent are always kept. The runner executes the text inside the first
\texttt{<action>} tag of a reply, or the rest of the reply if the tag is never
closed. On ALFWorld and WebShop it can fall back to the reply's last line; no other
text reaches the environment. A reply without an action gets a fixed
one-sentence reminder, and on the benchmarks whose actions are code, the first
three such replies of an episode do not use up a turn.

\newpage 
\subsection{Processors and knobs}
\label{app:method:interface}
A processor is a Python function \texttt{advise(ctx)} that returns a
dictionary. Its context exposes the turn index and budget, the task, the last
observation and action, the available actions, the steps taken so far, and a
scratchpad shared by the processors of one episode, but never the model's
reasoning or the reply being processed. Each hook accepts one key:
\begin{itemize}
\item \texttt{before\_model} may return \texttt{inject}, a message that only the
  next model call sees;
\item \texttt{after\_model} may return \texttt{rewrite\_action}, which replaces
  the extracted action;
\item \texttt{after\_step} may return \texttt{rewrite\_obs}, which replaces the
  observation the model reads. The trajectory record that the analyst reads keeps the environment's original text.
\end{itemize}
Processors on one hook run in a fixed order, each with an allow-listed set of
built-ins and the modules \texttt{re} and \texttt{json} but no imports, and an
exception inside a processor is caught and counts as an empty return.
Table~\ref{tab:knobs} lists the five knobs, and every composition sets all
five.

\begin{table}[H]
\caption{The five runner knobs, their values in the neutral harness, and their effect.}
\label{tab:knobs}
\begin{center}
\small
\begin{tabular}{@{}llp{0.58\linewidth}@{}}
\toprule
Knob & Neutral & Effect \\
\midrule
Protocol & tagged & tagged text replies, or one generic \texttt{act} tool call \\
Available actions & shown & append the environment's list of legal actions to each observation \\
Observation cap & off & truncate long observations (command actions only) \\
Action fallback & on & without a tag, take the reply's last line as the action (command actions only) \\
Remaining turns & hidden & prefix each observation with the number of turns left \\
\bottomrule
\end{tabular}
\end{center}
\end{table}

The remaining-turns knob gives the agent information that the neutral harness
withholds, and since the harnesses \method{} returned turn it on, part of any
gain over the neutral harness can come from information that the composer
granted rather than from what the loop learned.

\newpage 
\subsection{Components and assembly}
\label{app:method:assembly}
A component has an identifier, a kind, a title, a one-line summary, a body, an
applicability claim, a rationale and usage statistics; a processor also has a
hook and an order. Only the body can reach a harness, and
Table~\ref{tab:kinds} lists the five kinds. The librarian and the composer
read the library through an index with one line per component, which gives
the component's kind, summary and applicability claim, the rounds in which it
was deployed, the mean train score of those rounds, and how often it fired.
Flags mark components that were deployed but never fired and components never
tried, and the bodies follow the index, each cut at 1{,}600 characters. The
applicability claim uses seven fixed axes, such as the bottleneck, the horizon
and the density of the reward; the meta-agents read it, but no check uses it.

\begin{table}[H]
\caption{Component kinds, where the assembly places each, and when it acts.}
\label{tab:kinds}
\begin{center}
\small
\begin{tabular}{@{}p{0.15\linewidth}p{0.25\linewidth}p{0.28\linewidth}p{0.20\linewidth}@{}}
\toprule
Kind & Body & Placed & Acts \\
\midrule
Prompt fragment & an instruction or a worked example & system prompt, after the frame, in its order & every model call \\
Knowledge & facts observed in trajectories & system prompt, after the fragments & every model call \\
Processor & a Python function \texttt{advise(ctx)} & one runner hook & each turn it returns a key \\
Knob & runner settings & read by the composer only & never directly \\
Strategy & advice on composing & the composer's and librarian's input & never reaches the agent \\
\bottomrule
\end{tabular}
\end{center}
\end{table}

Given the library and a composition, the fixed assembly $\mathcal{B}$ builds
the harness in four steps. It first resolves and admits the selection: each
selected identifier is looked up, an identifier merged away is followed to the
component that absorbed it, every body passes the identity filter again,
processors must compile and pass a smoke test, and knob bodies must parse. It
then writes the system prompt, with the frame first, the selected prompt
fragments in their order, the knowledge bodies, and the runner's reply-format
block of 198 characters (381 where actions are code), all joined by blank
lines; titles, summaries and rationales are never placed. Finally, it compiles
each selected processor onto its hook and applies the composition's five knob
values last, so a knob component informs the composer but never sets a value
itself.

\newpage 
\subsection{The meta-agents}
\label{app:method:seats}
Each meta-agent receives its fixed instructions as the system message and one user
message that the loop assembles, and it returns JSON that follows a fixed
schema, with no tools or file access. The analyst is Claude Sonnet~5 and the
editor, librarian and composer are Claude Opus~4.6, all called with high
effort, adaptive thinking and no sampling parameters. Output budgets are
32{,}000 tokens for the analyst and the librarian and 16{,}000 for the others,
and a truncated reply is retried with twice the budget.
Table~\ref{tab:seats} lists what each meta-agent reads and returns, and
Appendix~\ref{app:method:prompts} gives their instructions in full.

\begin{table}[H]
\caption{What each meta-agent reads and returns in one round, in call order. Instructions are omitted.}
\label{tab:seats}
\begin{center}
\small
\begin{tabular}{@{}p{0.15\linewidth}p{0.45\linewidth}p{0.34\linewidth}@{}}
\toprule
Meta-agent (model) & Reads & Returns \\
\midrule
Analyst\newline (Sonnet~5) & the running harness's settings and full system prompt; outcome statistics over the round; 13 task groups with all $M$ trajectories each & a task signature; failure modes with their share, evidence and mechanism; candidate components, with a cited step for each fact \\
Editor\newline (Opus~4.6) & the analyst's output with its citation checks; a journal of earlier rounds (scores, decisions, library edits, rejected processors) & the selected candidates, rewritten or merged; dropped candidates with reasons; then the processor check \\
Librarian\newline (Opus~4.6) & the library index with usage statistics and all bodies; the editor's selections and drops & operations: add, update or merge a complete component, or drop one, each with a reason \\
Composer\newline (Opus~4.6) & the task signature; this round's score and the best harness's; the strategies; the library; the journal; the best harness's selection, knobs and size (not its frame); the editor's picks; a 4{,}000-character excerpt of failed trajectories & a selection of components, a frame and five knob values \\
\bottomrule
\end{tabular}
\end{center}
\end{table}

\paragraph{Strategies.}
The three strategies we write are generic rules for composing. The first says
not to write a reply-format paragraph, since the runner appends its own, and
the second tells the composer to write the frame and let components carry the
content. The third says that dropping a component that was live in the best
round is a decision the composer must name in its rationale.

\newpage 
\FloatBarrier
\subsection{Sampling, checks and filters}
\label{app:method:checks}

\paragraph{Failure-Oriented Group Sampling.}
Algorithm~\ref{alg:groupsample} gives the rule. Only tasks that all $M$
workers attempted are eligible, and the sample holds the ten mixed groups with
the highest mean partial credit and the three all-fail groups with the lowest.
If fewer than ten mixed groups exist, all-fail groups with the highest partial
credit fill the gap.

\begin{algorithm}[H]
\caption{\textsc{GroupSample}, the analyst's failure-oriented group sampling. $n_{\mathrm{mix}}=10$ and $n_{\mathrm{fail}}=3$, or 5 and 2 on SWE-bench Verified.}
\label{alg:groupsample}
\begin{algorithmic}[1]
\Require groups $\mathcal{G}$: for each train task $\tau$ that all $M$ workers attempted, its $M$ rollouts
\State for each $\tau$: $w_\tau\gets$ number of workers that solve $\tau$;\quad $\bar f_\tau\gets$ mean partial credit of its $M$ rollouts
\State $\mathcal{X}\gets\{\tau: 0<w_\tau<M\}$ sorted by $\bar f_\tau$, highest first \Comment{mixed}
\State $\mathcal{Z}\gets\{\tau: w_\tau=0\}$ \Comment{all-fail}
\State $S_{\mathrm{mix}}\gets$ the first $n_{\mathrm{mix}}$ tasks of $\mathcal{X}$
\State $S_{\mathrm{fail}}\gets$ the $n_{\mathrm{fail}}$ tasks of $\mathcal{Z}$ with the lowest $\bar f_\tau$
\If{$|S_{\mathrm{mix}}|<n_{\mathrm{mix}}$} \Comment{too few mixed groups}
  \State add to $S_{\mathrm{fail}}$ the $n_{\mathrm{mix}}-|S_{\mathrm{mix}}|$ remaining tasks of $\mathcal{Z}$ with the highest $\bar f_\tau$
\EndIf
\State \Return the groups of $S_{\mathrm{mix}}\cup S_{\mathrm{fail}}$, each with its $M$ rollouts and outcomes
\end{algorithmic}
\end{algorithm}

 Every group shows
all $M$ trajectories, so the analyst reads 52 trajectories when $M=4$, each
with its outcome, partial credit, number of turns and exit reason. Actions are
cut to 400 characters and observations to 300, with longer allowances for the
first turn and the five longest steps, and if the sample still exceeds 1.6
million characters, the allowances shrink until it fits, without dropping a
trajectory.

\paragraph{The Editor's Processor Check.}
Each processor the editor keeps is compiled, smoke-tested and replayed on the
first 24 episodes of the round, and a processor that fails to compile, raises
an error or never fires is dropped. The editor then reads each surviving
processor's claim, its code, how often it fired and up to six of its outputs,
dropping any processor whose outputs contradict its claim. Only processors face this check; fragments, knowledge and knobs reach the librarian without a replay.

\paragraph{The Identity Filter.}
One filter guards every meta-agent's input and output. Before any meta-agent reads a
trajectory, task identifiers, benchmark and split fields, and URLs are removed,
names that identify a benchmark in environment text are replaced by a
placeholder, and the turn cap appears only as a label, short or long, that at least two
benchmarks share. When the filter fires on an output, the response depends on the
meta-agent: an analysis is retried once and then discarded, editor candidates are
dropped, library writes are refused, and a frame is anonymized. Component bodies pass the filter again at assembly.

\paragraph{Citations for Facts.}
Each knowledge component that the analyst proposes cites the episode and turn
where the agent observed its fact, unless no single step yields it, as for a
regularity across episodes. The loop checks that the cited turn exists and
that the action there matches the fact, and it drops the component otherwise,
so that the analyst cannot store a fact recalled from pre-training as if the
agent had seen it.

\FloatBarrier
\subsection{The meta-agents' instructions}
\label{app:method:prompts}
\begin{figure}[H]
  \centering
  \includegraphics[width=\textwidth,height=0.9\textheight,keepaspectratio]{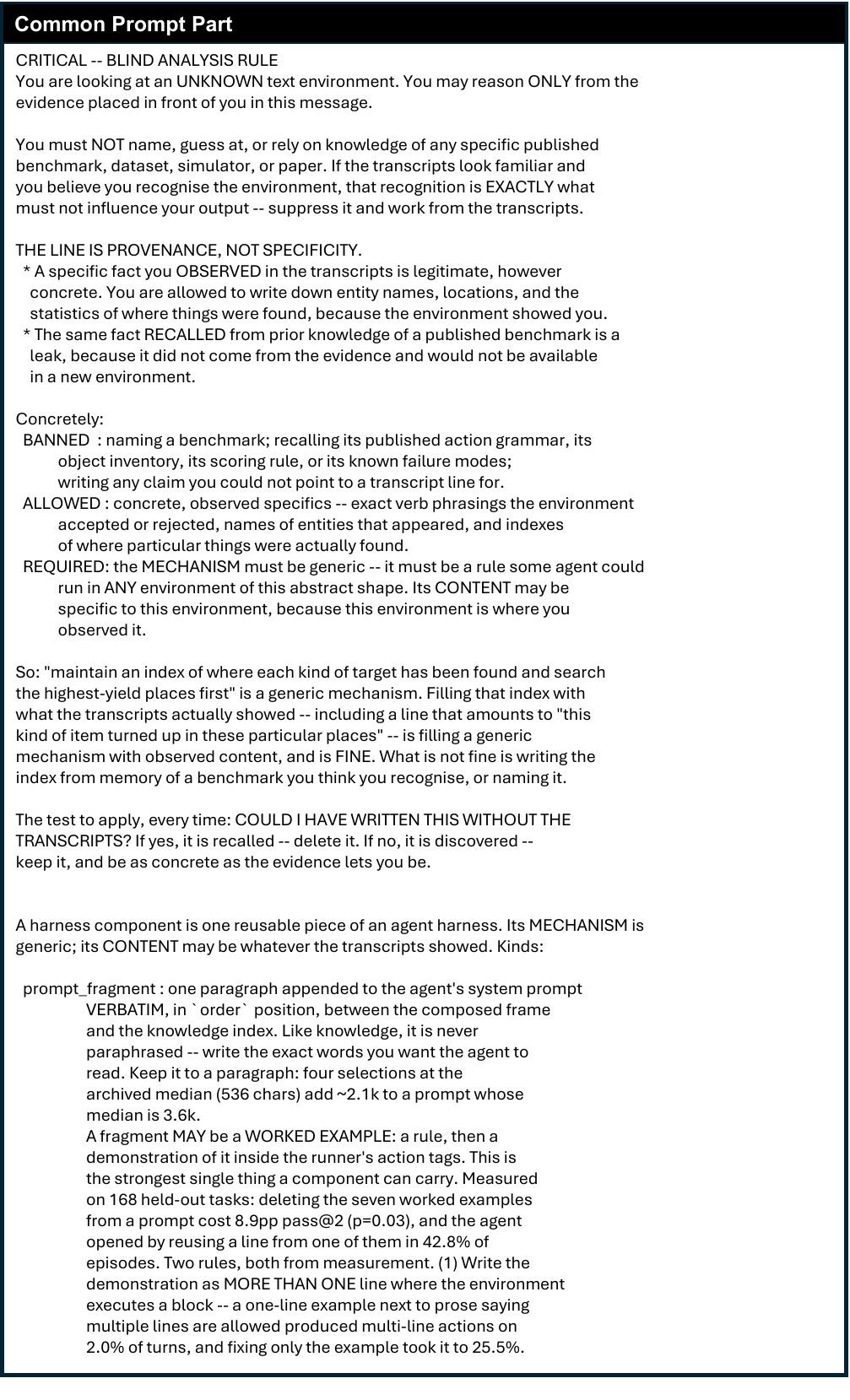}
  \caption{The part of the instructions that all four meta-agents share (1 of 4).}
  \label{fig:prompt-common-1}
\end{figure}

\begin{figure}[H]
  \centering
  \includegraphics[width=\textwidth,height=0.9\textheight,keepaspectratio]{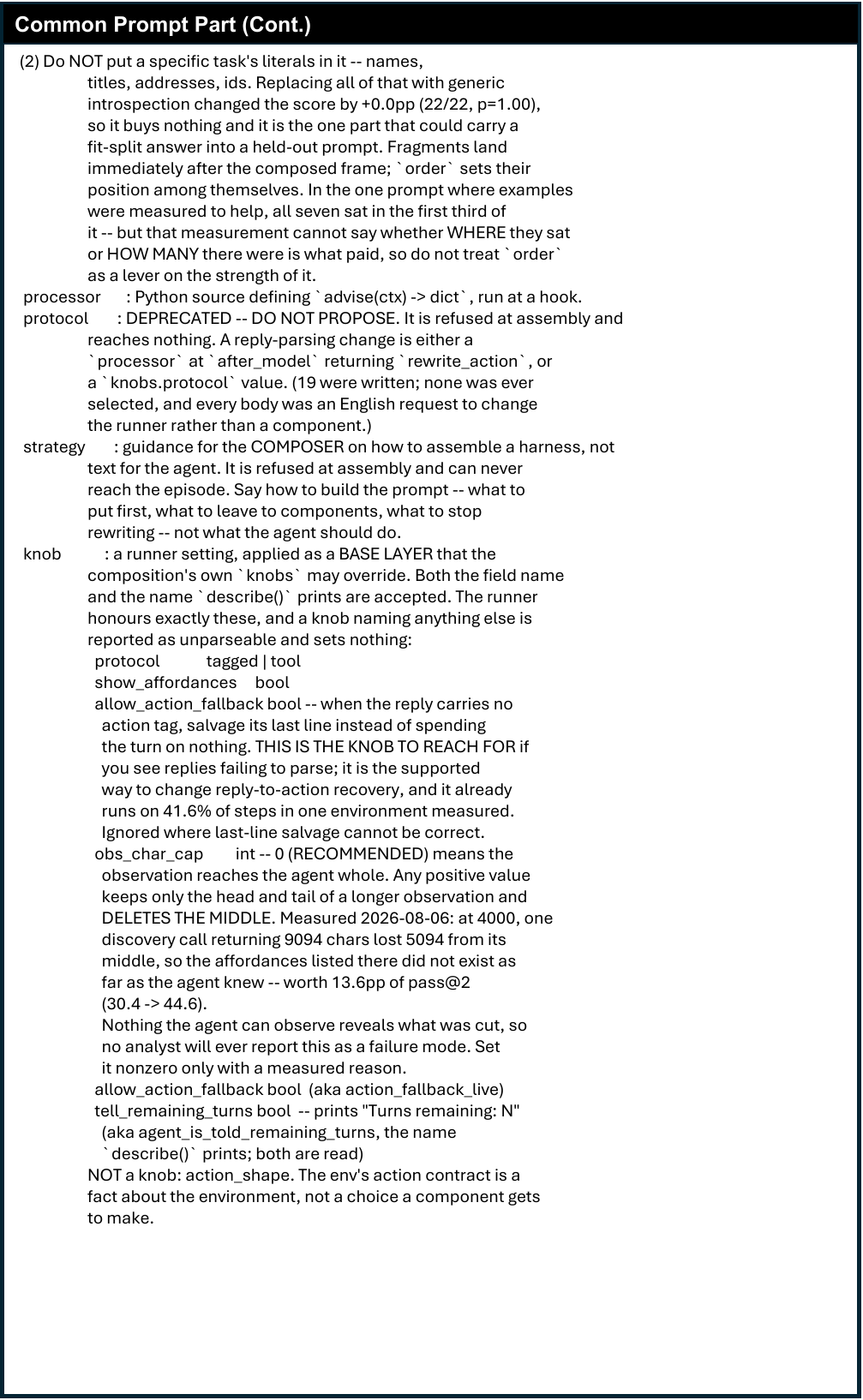}
  \caption{The part of the instructions that all four meta-agents share (2 of 4).}
  \label{fig:prompt-common-2}
\end{figure}

\begin{figure}[H]
  \centering
  \includegraphics[width=\textwidth,height=0.9\textheight,keepaspectratio]{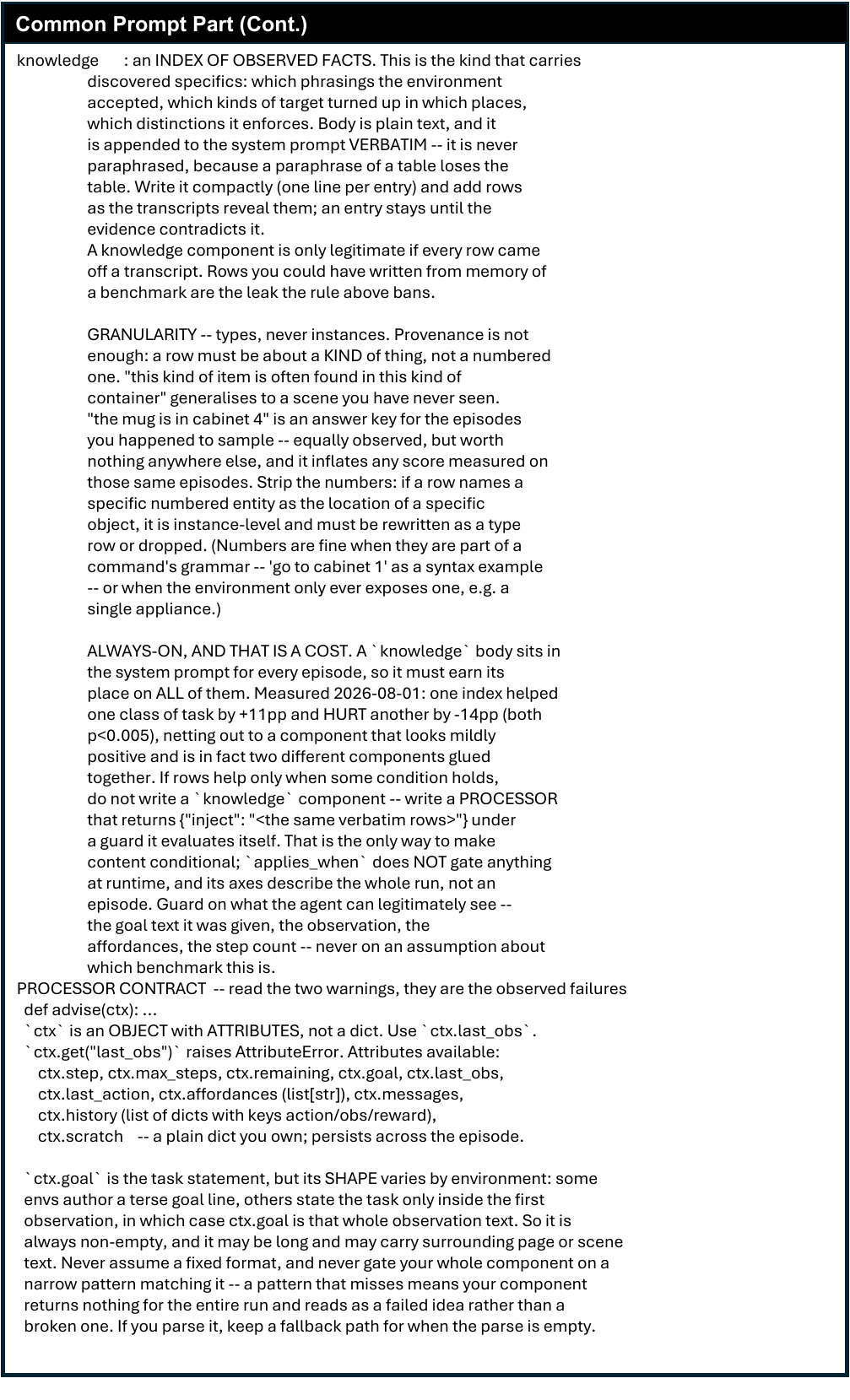}
  \caption{The part of the instructions that all four meta-agents share (3 of 4).}
  \label{fig:prompt-common-3}
\end{figure}

\begin{figure}[H]
  \centering
  \includegraphics[width=\textwidth,height=0.9\textheight,keepaspectratio]{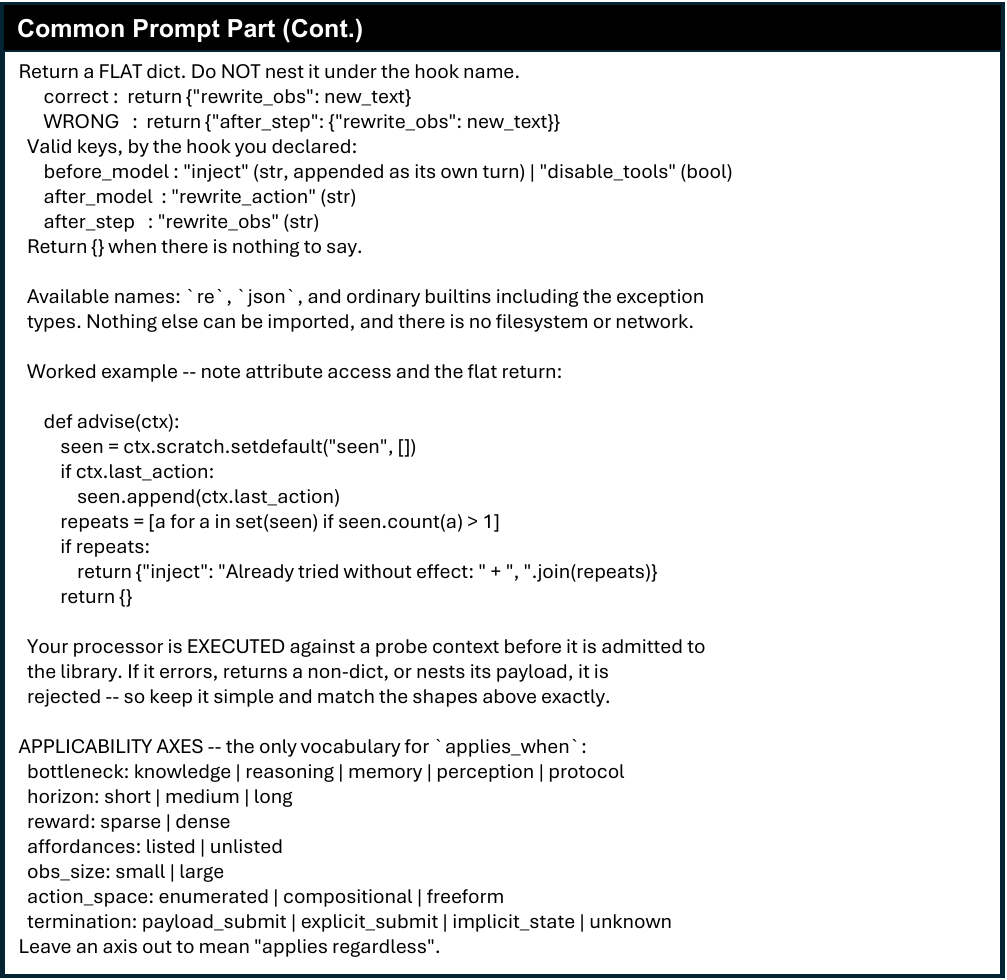}
  \caption{The part of the instructions that all four meta-agents share (4 of 4).}
  \label{fig:prompt-common-4}
\end{figure}

\begin{figure}[H]
  \centering
  \includegraphics[width=\textwidth,height=0.9\textheight,keepaspectratio]{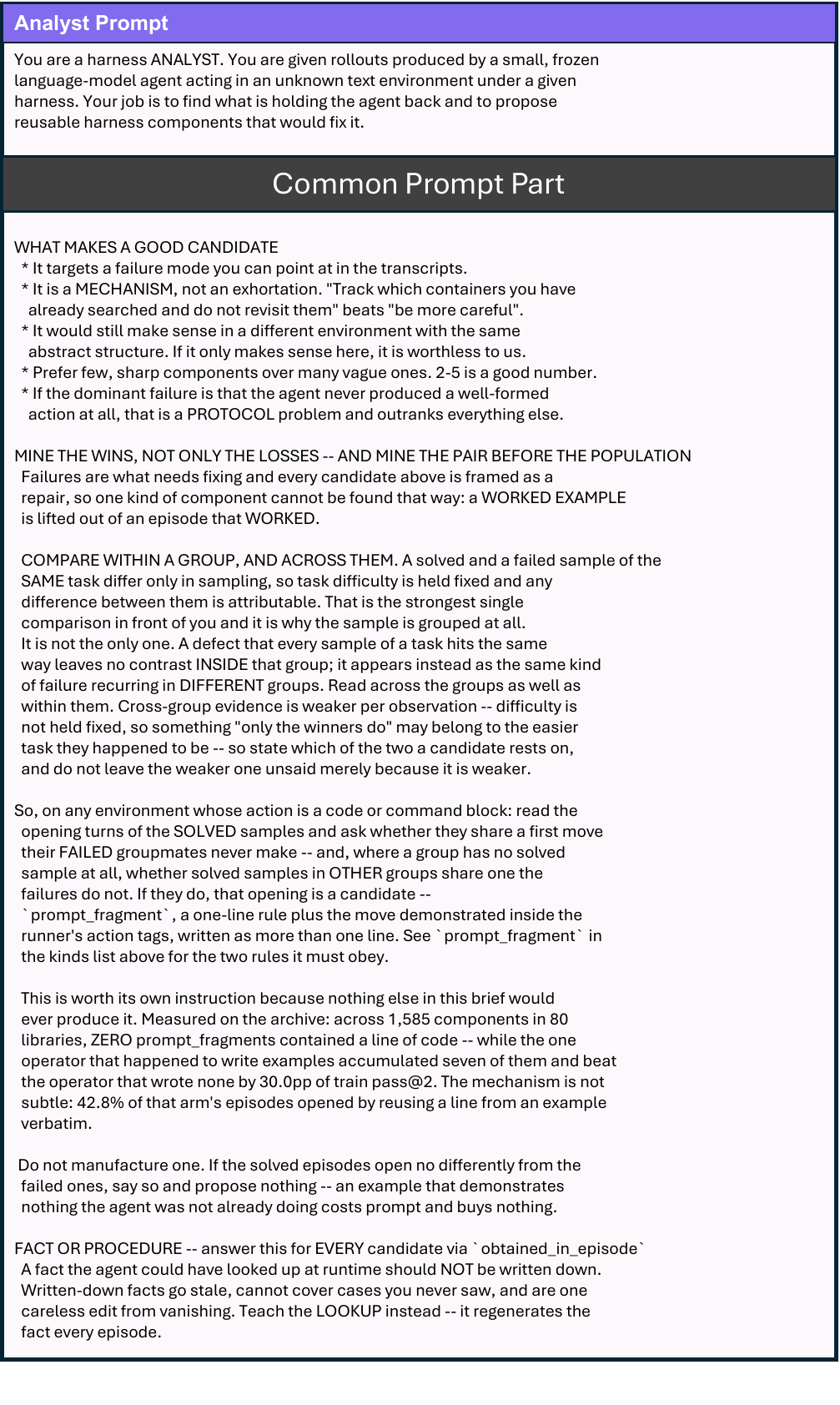}
  \caption{The analyst's instructions (1 of 2).}
  \label{fig:prompt-analyst-1}
\end{figure}

\begin{figure}[H]
  \centering
  \includegraphics[width=\textwidth,height=0.9\textheight,keepaspectratio]{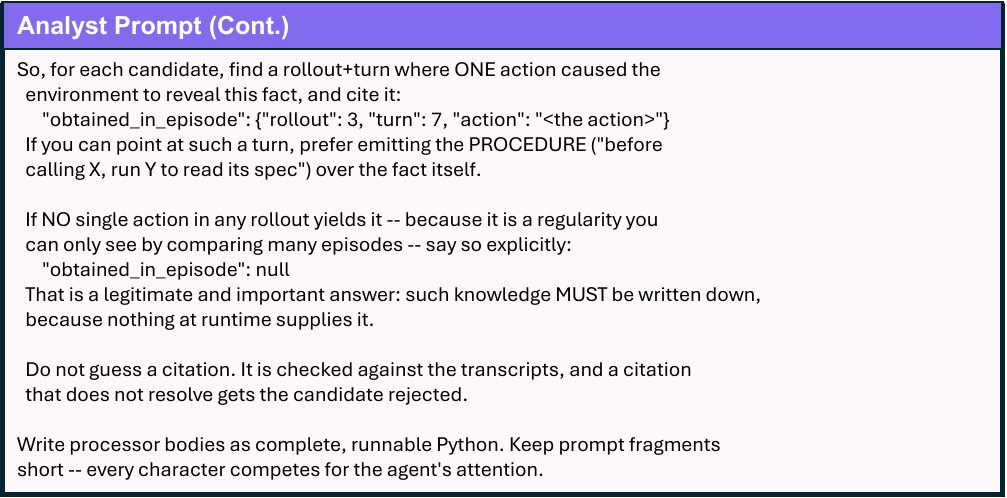}
  \caption{The analyst's instructions (2 of 2).}
  \label{fig:prompt-analyst-2}
\end{figure}

\begin{figure}[H]
  \centering
  \includegraphics[width=\textwidth,height=0.9\textheight,keepaspectratio]{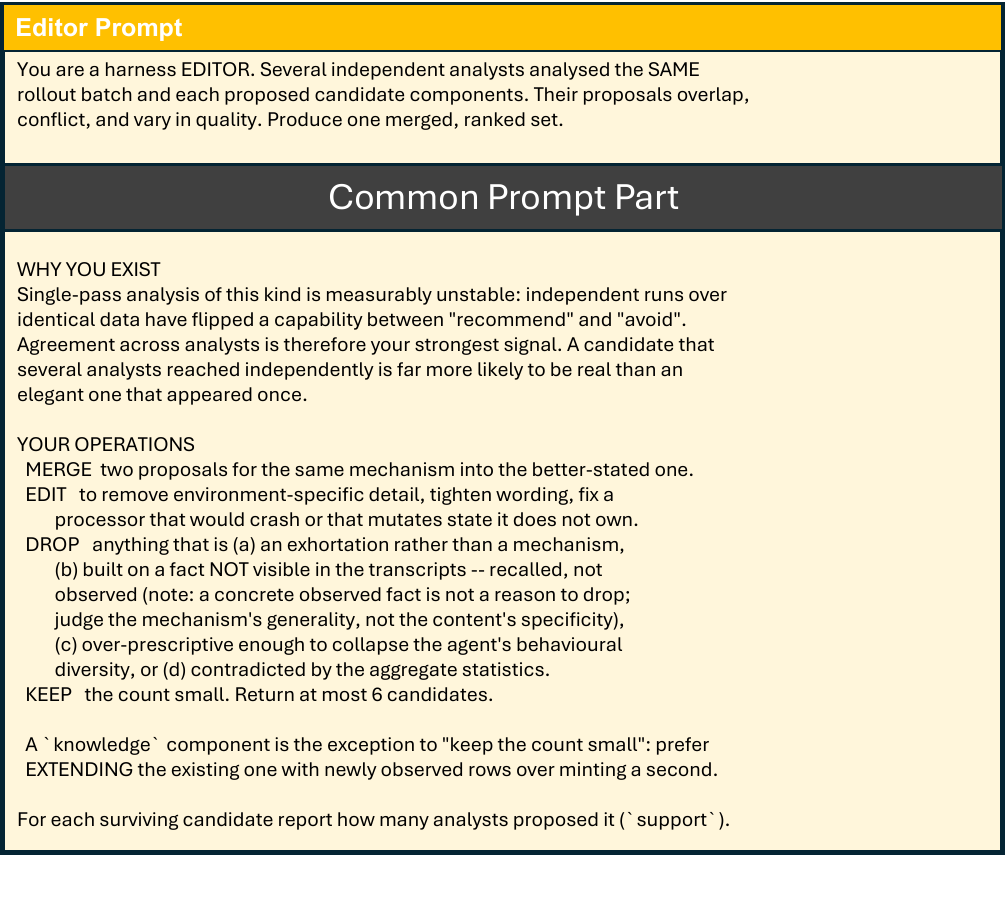}
  \caption{The editor's instructions.}
  \label{fig:prompt-editor}
\end{figure}

\begin{figure}[H]
  \centering
  \includegraphics[width=\textwidth,height=0.9\textheight,keepaspectratio]{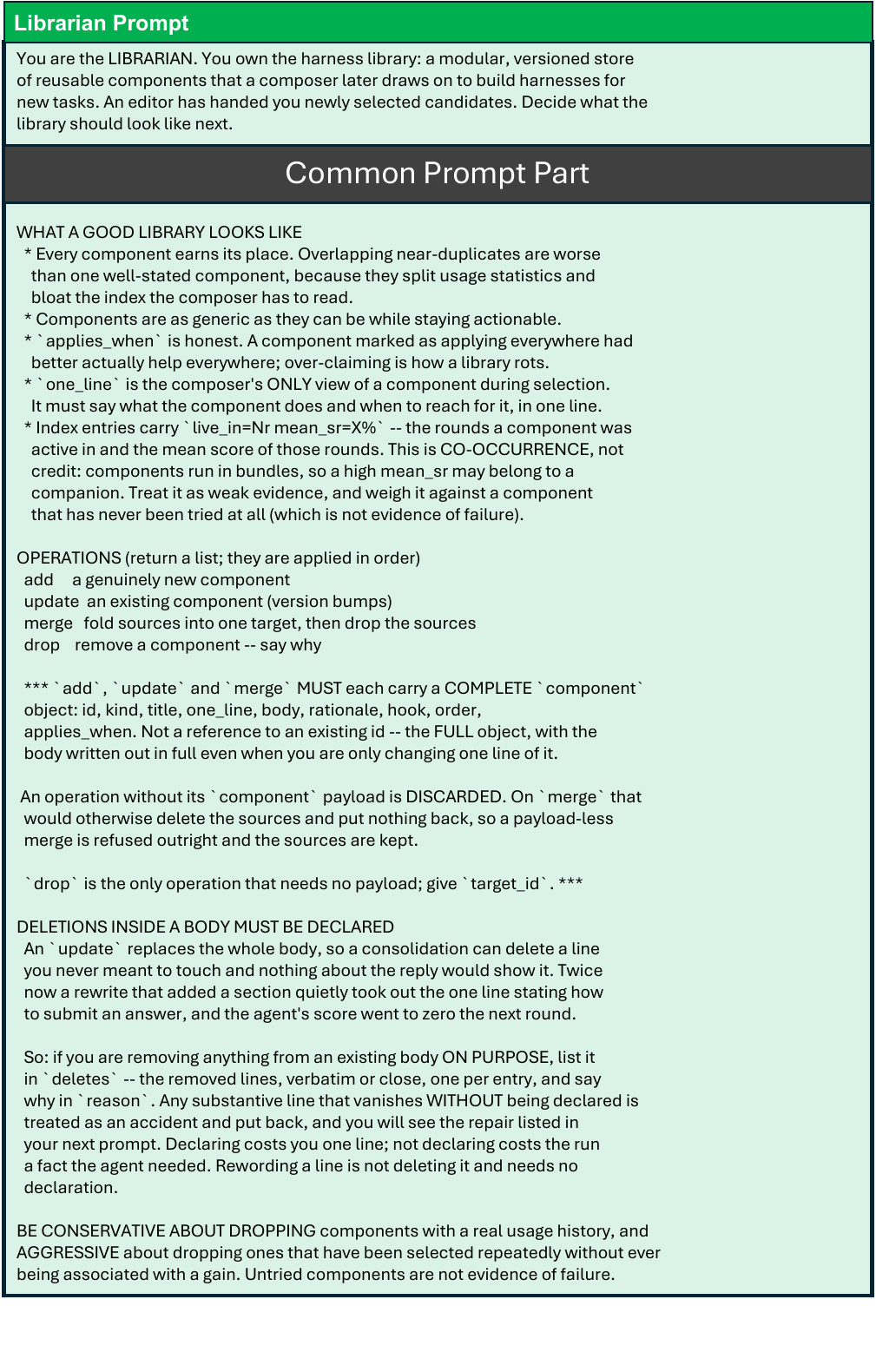}
  \caption{The librarian's instructions.}
  \label{fig:prompt-librarian}
\end{figure}

\begin{figure}[H]
  \centering
  \includegraphics[width=\textwidth,height=0.9\textheight,keepaspectratio]{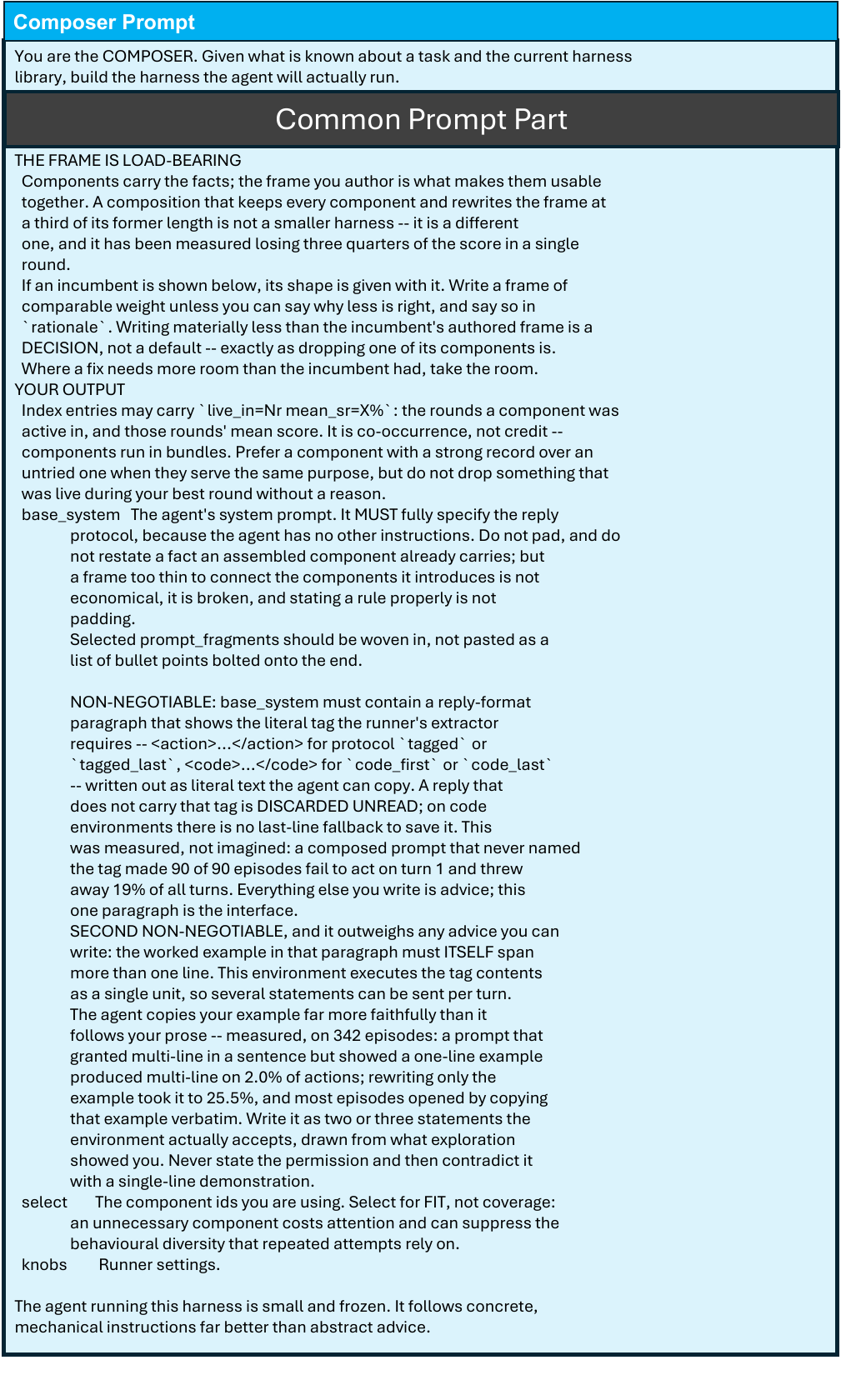}
  \caption{The composer's instructions.}
  \label{fig:prompt-composer}
\end{figure}

Figures~\ref{fig:prompt-common-1} to~\ref{fig:prompt-composer} reproduce the
meta-agents' instructions verbatim; the composer's instructions call the best
harness the incumbent. All four meta-agents share one part
(Figures~\ref{fig:prompt-common-1}--\ref{fig:prompt-common-4}), which holds
the rule against using recalled knowledge of a benchmark and documents the
component kinds, the processor interface and the applicability axes. Each
meta-agent's own instructions mark where this part is inserted. The loop adds one user message per call (Table~\ref{tab:seats}).


\subsection{The gate}
\label{app:method:gate}
Let $v_{j,i}$ be the pass@1 of worker $i$ on the val tasks in round $j$, and
let $M'$ be the number of val rollouts per task: $M'=4$ in the inductive runs, and
$M'=M$ in the transductive runs, which reuse the train rollouts. The
noise band after round $r$ pools the within-round spread of these scores,
\begin{equation}
  n_r \;=\; 2\,\hat\sigma_r\sqrt{2/M'},\qquad
  \hat\sigma_r=\frac{1}{r}\sum_{j\le r}\frac{\max_i v_{j,i}-\min_i v_{j,i}}{d_2(M')},
\end{equation}
with the range-to-deviation constants $d_2(2)=1.128$ and $d_2(4)=2.059$. The
margin is $\delta=\max\big(\min(7,\,b^\star/2),\ n_r,\ \Delta^\star\big)$
pp, where $\Delta^\star$ is the spread of the best harness's repeated
measurements, and the loop applies Eq.~(\ref{eq:gate}) by checking the
following conditions in order.
\begin{enumerate}
\item In the first round, the harness that ran becomes the best harness.
\item If a revert made the best harness run again, the loop holds, and $b^\star$ becomes the mean of its measurements.
\item If more than 5\% of its replies were unparsed, the loop holds.
\item If $s_t>b^\star$, it accepts.
\item If $s_t<b^\star-\delta$, it reverts.
\item Otherwise it holds.
\end{enumerate}
The loop scores a harness in the round after its composer wrote it, from the
rollouts that the analyst of that round also reads, so the composition written
in the last round is never scored. A revert discards the composition built
from the reverted harness's rollouts, while the library keeps that round's
edits.

\paragraph{Output and Grading.}
The output is the last accepted best harness, whose system prompt is rebuilt
from the library snapshot of the round that composed it and must match the
original byte for byte. A test run then gives it two fresh attempts per test
task, and with two attempts the per-task union equals the estimator of
Eq.~(\ref{eq:passk}).

\subsection{Human-written inputs}
\label{app:method:priors}
This is the complete list of human-written inputs; none of them names a
benchmark, and the meta-agents' instructions pass the same identity filter
that guards a run.
\begin{enumerate}
\item Text for the agent: the neutral harness, the runner's
  reply-format block, the runner's labels (task, observation, available
  actions, turns remaining), and the reminder sent after a reply without an
  action.
\item Instructions for the meta-agents: a role description for each,
  the rule against using recalled knowledge of a benchmark, documentation
  of the component kinds, the processor interface and the applicability axes,
  and a few measurements from our development runs, stated without naming an
  environment; some came from AppWorld, including the split of its inductive
  results (Appendix~\ref{app:method:prompts}).
\item The three strategies (Appendix~\ref{app:method:seats}).
\item The workflow: the order of the stages, the sampling rule, the
  admission checks, the gate, $R$ and $M$.
\item Each benchmark's interface: whether actions are code or
  commands, the turn cap, the generation budget and, on WebShop, the syntax of
  its two actions (Appendix~\ref{app:benchmarks:list}).
\end{enumerate}

\newpage 
\subsection{Configuration}
\label{app:method:config}
Table~\ref{tab:config} lists the settings of all runs.

\begin{table}[H]
\caption{Settings of the \method{} runs. Exceptions: SWE-bench Verified uses $M=2$, five mixed plus two all-fail groups and two val rollouts per task.}
\label{tab:config}
\begin{center}
\small
\begin{tabular}{@{}p{0.27\linewidth}p{0.67\linewidth}@{}}
\toprule
Setting & Value \\
\midrule
Agent & Qwen3.5-9B in thinking mode; temperature 1.0, top-$p$ 0.95, top-$k$ 20, presence penalty 1.5 \\
Generation and context & 4{,}096 tokens per reply; context window 40{,}960 tokens; prompt budget 32{,}768 tokens \\
Turn caps & 20, 20, 40, 20 and 200 (ALFWorld, WebShop, AppWorld, GAIA, SWE-bench Verified) \\
Rounds and workers & $R=8$ (round~0 and seven evolution rounds), $M=4$ \\
Val rollouts & 4 per task (inductive); the train rollouts (transductive) \\
Analyst sample & 10 mixed and 3 all-fail task groups, each with all $M$ trajectories \\
Editor & at most six candidates; journal of earlier rounds; processor check on 24 episodes \\
Gate & pass@2; accept if $s_t>b^\star$; revert if $s_t<b^\star-\delta$ \\
Grading & three test runs of two attempts per task; the harness is rebuilt byte for byte \\
Meta-agents & analyst Claude Sonnet~5; editor, librarian and composer Claude Opus~4.6; high effort \\
\bottomrule
\end{tabular}
\end{center}
\end{table}

\clearpage
\section{Benchmarks and evaluation}
\label{app:benchmarks}

\subsection{Benchmarks}
\label{app:benchmarks:list}
Table~\ref{tab:benchmarks} lists the task sets, whose splits are disjoint from
one another at the task level, and the adapters pass the environments' own text
to the agent.

\begin{table}[H]
\caption{Task sets. Inductive runs evolve on the train tasks, gate on the val tasks and grade on the test tasks. Transductive runs use the test tasks for all three.}
\label{tab:benchmarks}
\begin{center}
\small
\begin{tabular}{@{}lcccc@{}}
\toprule
Benchmark & Turn cap & Train & Val & Test \\
\midrule
ALFWorld & 20 & 134 (train) & 140 (valid-seen) & 134 (valid-unseen) \\
WebShop & 20 & 100 & 100 & 100 \\
AppWorld & 40 & 90 (train) & 57 (dev) & 168 (test-normal) \\
GAIA & 20 & -- & -- & 103 (transductive only) \\
SWE-bench Verified & 200 & 55 & 55 & 55 \\
\bottomrule
\end{tabular}
\end{center}
\end{table}

\paragraph{ALFWorld} casts six household task types as text games. The agent
receives the game's own opening text, which states the goal, and each
observation lists the commands the environment admits; an action is one
command, such as ``heat apple 1 with microwave 1''. The episode ends when the
goal is met or after 20 turns, with no partial credit, and the test games are
set in rooms that no training game uses.

\paragraph{WebShop} asks the agent to buy a product that matches an
instruction, on a simulated shop with 1{,}000 products and an in-memory BM25
search. The environment accepts \texttt{search[}query\texttt{]} and
\texttt{click[}label\texttt{]} but never states this syntax, so our adapter
lists the legal actions in that form, the only action syntax we supply on any
benchmark. Buying ends the episode with a reward in $[0,1]$ for how well the
product matches, and success is a reward of 1; the three splits are windows of
100 instructions with no product in common.

\paragraph{AppWorld} simulates nine everyday apps with 457 APIs. The agent sees
only the task and the identity of the user it acts for, and since nothing
states that actions are Python or that the APIs exist, it must discover both
from tracebacks and from the environment's documentation app. The episode ends
when the agent calls the API that declares the task complete, or after 40
turns, and state-based unit tests then score it: partial credit is the fraction
of tests passed, and success requires all of them. We use the official train,
dev and test-normal splits and report the harder test-challenge split separately
(Appendix~\ref{app:aw-challenge}).

\paragraph{GAIA} poses questions that need web research, files and code, each
with one reference answer. Since GAIA ships no environment, our bridge
provides one, whose opening text defines six actions after the question:
search, open a page, read a local file, run Python, run a shell command, both
without network access, and submit. Submitting ends the episode, and the answer
scores 1 if it matches the reference under GAIA's normalized exact match. We use
the 103 text-only validation questions on which HarnessX also reports; since
their answers are public, the test answers cannot be withheld, and 103
questions are too few to split three ways, so GAIA is transductive only.

\paragraph{SWE-bench Verified} asks the agent to resolve a GitHub issue by
editing the repository, and each episode runs in the official container of its
instance with networking disabled. The agent sees the issue and a listing of
the repository, and each action is a shell command. The episode ends when the
agent submits or after 200 turns, and the official harness then grades the
resulting patch with tests whose names the agent never sees.
After excluding three instances, we draw three disjoint sets of 55
instances, stratified by repository and difficulty.

\subsection{Grading}
\label{app:benchmarks:grading}
With $n\ge k$ attempts per task, $c_\tau$ of which succeed on task $\tau$, we
estimate pass@$k$ with the unbiased estimator of \citet{chen2021codex}:
\begin{equation}
  \mathrm{pass@}k(h;\mathcal{T})
  =\frac{1}{|\mathcal{T}|}\sum_{\tau\in\mathcal{T}}
  \left[1-\binom{n-c_\tau}{k}\Big/\binom{n}{k}\right].
  \label{eq:passk}
\end{equation}
A test run gives the returned harness two fresh attempts per test task: its
pass@1 is the mean success rate of the two attempts, and its pass@2 is the
share of tasks solved by either, which equals Eq.~(\ref{eq:passk}) with
$n=k=2$. Tables average three test runs, and episodes lost to infrastructure
faults are excluded from both counts. The loop's own scores, such as the curves of
Figure~\ref{fig:appworld-evolution}, use Eq.~(\ref{eq:passk}) over the $M$ rollouts of a round.
The SWE-bench Verified inductive run differs from the others in two settings.
Its analyst also receives the share of episodes that never edited a file, a
statistic we added after inspecting test-split episodes of an earlier
SWE-bench Verified run, and knobs that the runner cannot honour for the
benchmark's action format are dropped rather than composed.

\subsection{API cost}
\label{app:benchmarks:cost}
The API cost counts only the meta-agents' tokens in the calls that produce $h_1$
to $h_7$: the last round's calls compose a candidate that is never run, and we
do not count them. Since the agent runs on local GPUs, neither method's rollouts
cost API calls. For HarnessX we sum the token counts of every call in its first
seven evolution steps, context compaction included. For \method{}, whose logs
record each meta-agent's cumulative token counts after every round, we take the
cost up to the composition of $h_7$, and where a run's log is lost, seven
eighths of the run's total. We price both at \$5 and \$25 per million input
and output tokens for Opus~4.6 and at the introductory \$2 and \$10 for
Sonnet~5. Both pipelines cached their fixed
system prompts during the runs, but we bill every input token at the full
price, cached or not, so that no cost depends on how a run used the cache.
For five runs whose token records were lost, the SWE-bench Verified
transductive run of \method{} and four Qwen3.5-4B runs (ALFWorld in both
regimes, GAIA and SWE-bench Verified transductive), only the total recorded for
the run survives.
We convert it to our convention with the ratio measured on a run of the same
benchmark whose records survive and take seven eighths of the result.

Table~\ref{tab:main_results}'s cost gap has two sources: the number of calls,
and the analyst's cheaper model, which reads most of \method{}'s tokens. HarnessX's meta-agent
made 65 to 130 Opus calls per round, because each round is an agentic session
of dozens of turns, whereas \method{} made about five calls per round on the
four benchmarks whose records survive. HarnessX's calls averaged
43{,}000 to 50{,}000 input tokens, while \method{}'s analyst read more per
call, 65{,}000 to 365{,}000, and its other meta-agents far less, 6{,}000 to
12{,}000, so that HarnessX consumed 3.3 to 5.6 million input tokens per round
against 0.12 to 0.54 million for \method{}. The trajectory text itself is of
comparable size: about 38{,}000 to 217{,}000 tokens per round reach HarnessX
through its tools, and at most 57{,}000 to 357{,}000 reach \method{}'s analyst
per call. HarnessX, however, keeps what it reads in a context that it re-sends
on every later call, which accounts for 25\% to 36\% of its input.

\newpage 
\subsection{Answer copies on GAIA}
\label{app:benchmarks:gaia}
GAIA's questions and answers are public, and copies of them sit on the open
web, in dataset pages, papers and code repositories. A searching agent can
reach a copy in two ways: it can open a page that states the answer, or it can
search the question's text verbatim, which tends to surface such pages.
Because the web changes, no two runs see the same pages, and a GAIA result
cannot be reproduced exactly. \method{}'s search and page tools block known
copies of the benchmark, namely pages on a list of known mirrors, results that
contain a task's answer record or its annotator metadata, and GAIA task
identifiers, and they fail closed.

Mirrors missing from the list are not blocked: in the \method{}
run that Table~\ref{tab:main_results} reports, a list-free scan of observations found 30 cases
in which the answer reached the agent from such a copy before the agent had
produced it. In the round that ran the returned harness, 10 of these cases
fell in winning episodes. Scoring those episodes as failures would leave that round's pass@2 no
lower than 62.9, against 64.6 as measured, an upper bound on what the copies
bought, since the agent may also have found the answer on its own. Evolution also
learned from these episodes: one returned processor tells the agent that hard-to-find facts sometimes appear on dataset mirror pages. We report the run's scores as
measured; the scan covered the loop's episodes only.

\subsection{The HarnessX reproduction}
\label{app:benchmarks:harnessx}
The released HarnessX code ships its harness framework, a single-agent
meta-agent, adapters for GAIA and SWE-bench, and an evolution recipe for GAIA,
but not the four-stage evolution engine on which the paper's results
rest~\citep{harnessx}, so we run the released single-agent meta-agent on every
benchmark. In the paper's own comparison on GAIA, a single-agent evolver scores 1.0~pp
lower than the four-stage engine and uses about 14\% more tokens.
Table~\ref{tab:harnessx-provenance} lists which parts of the reproduction are
official and which we wrote, and we cap evolution at eight rounds with one
run, allow 4{,}096 tokens per reply, and use our turn caps. On four
benchmarks, its meta-agent runs without extended thinking.
On WebShop, AppWorld, GAIA and SWE-bench Verified we also turned on HarnessX's
optional caching of its system prompt, which leaves its costs unchanged,
because Appendix~\ref{app:benchmarks:cost} bills every input token at the full
price.
HarnessX's loop scores pass@2 in every round and reports its best round on the evolution tasks, as the paper does.
On GAIA, HarnessX peaked at round~2, and we evaluate that round's harness.

\begin{table}[H]
\caption{Provenance of each part of our HarnessX runs. \emph{Official} is the released code, unmodified; \emph{patched} adds our changes; \emph{ours} is absent from the release and written by us. The meta-agent is Claude Opus~4.6 on every benchmark.}
\label{tab:harnessx-provenance}
\begin{center}
\small
\begin{tabular}{@{}lp{0.11\linewidth}p{0.15\linewidth}p{0.15\linewidth}p{0.23\linewidth}@{}}
\toprule
Benchmark & Seed harness & Evolution recipe & Meta-agent reasoning & Other deviations \\
\midrule
ALFWorld & ours & ours & thinking off & task seeds cover 88 of the 134 test games \\
WebShop & ours & ours & thinking off & BM25 search \\
AppWorld & ours & ours & thinking off & not in the HarnessX paper; no early stopping \\
GAIA & official & official, patched (pass@2) & official (extended thinking) & a different search API; answer judge off \\
SWE-bench Verified & official & ours & thinking off & 200 turns (release: 60); our grader \\
\bottomrule
\end{tabular}
\end{center}
\end{table}

Three differences favour one side or the other: (i)~HarnessX's released
SWE-bench runner, which we kept as released, names in the agent's task
description the tests that must pass, whereas \method{}'s agent never sees
them; (ii)~the task seeds of HarnessX's ALFWorld run cover only 88 of the 134
test games; and (iii)~AppWorld is not a HarnessX benchmark, so its HarnessX
seed is ours, written in the style of HarnessX's other seeds.

\clearpage
\section{Additional results}
\label{app:results}
Section~\ref{sec:results} gives each result below in one sentence; unless a caption or the text says otherwise, the agent is
Qwen3.5-9B, the runs are transductive, and scores average three test runs.

\FloatBarrier
\subsection{A smaller backbone: Qwen3.5-4B}
\label{app:4b}
Table~\ref{tab:qwen35_4b_evolite} repeats the main experiment with
Qwen3.5-4B, starting from the neutral harness. \method{} raises pass@2 in all
nine benchmark and regime cells by 6.7 to 49.8~pp, at \$7.78 to \$22.45
per run, with the largest gains on WebShop and ALFWorld and the smallest on
SWE-bench Verified. The inductive harnesses stay within 2.7~pp of the
transductive ones.

\begin{table}[H]
    \centering
    \caption{Test pass@2 (\%) of Qwen3.5-4B with the harness that \method{} evolves from the neutral harness, averaged over three test runs. \textit{Initial} is the neutral harness. $\Delta$ is Evolved $-$ Initial, before rounding. The GAIA run blocks known copies of the benchmark's answers.}
    \label{tab:qwen35_4b_evolite}
    \small
    \setlength{\tabcolsep}{7pt}
    \renewcommand{\arraystretch}{1.08}
    \begin{tabular}{llccccc}
        \toprule
        Benchmark & Mode
        & \multicolumn{2}{c}{pass@2 (\%)} & $\Delta$ & \shortstack{Best\\Round} & API cost (\$) \\
        \cmidrule(lr){3-4}
        & & Initial & Evolved & & & \\
        \midrule
        \multirow{2}{*}{ALFWorld}
            & Transductive & 44.8 & 88.3 & +43.5 & 7 & 7.78 \\
            & Inductive    & 44.8 & 87.3 & +42.5 & 7 & 7.86 \\
        \midrule
        \multirow{2}{*}{WebShop}
            & Transductive & 15.2 & 65.0 & +49.8 & 4 & 9.06 \\
            & Inductive    & 15.2 & 62.3 & +47.2 & 2 & 11.39 \\
        \midrule
        \multirow{2}{*}{AppWorld}
            & Transductive & 0.0  & 33.9 & +33.9 & 6 & 11.32 \\
            & Inductive    & 0.0  & 33.7 & +33.7 & 5 & 12.23 \\
        \midrule
        GAIA
            & Transductive & 33.7 & 55.3 & +21.7 & 7 & 10.64 \\
        \midrule
        \multirow{2}{*}{SWE-bench Verified}
            & Transductive & 46.0 & 53.9 & +7.9  & 3 & 15.31 \\
            & Inductive    & 46.0 & 52.7 & +6.7  & 7 & 22.45 \\
        \bottomrule
    \end{tabular}
\end{table}

\FloatBarrier
\subsection{AppWorld test-challenge split}
\label{app:aw-challenge}
Table~\ref{tab:qwen9b-appworld-challenge} grades the AppWorld harnesses on the
test-challenge split, whose 417 tasks need apps that the train split never
uses; the bare agent and the neutral harness solve none of them. \method{}'s
harnesses reach 48.6 pass@2 when evolved inductively and 49.2 when evolved on
test-normal, against 50.5 for HarnessX. HarnessX, however, starts from a
handcrafted AppWorld seed that alone scores 32.8, and its run costs \$190.32
against \$14.56 and \$12.51 for \method{}.

\providecommand{\method}{LiteEvo}
\begin{table}[H]
    \centering
    \caption{Results on AppWorld's test-challenge split, whose tasks need apps unseen in training (Qwen3.5-9B). \textit{Bare} gives the agent only the task text, and \textit{Bare + format} adds a minimal hint on the reply format. \textit{Neutral} is \method{}'s starting harness, and \textit{HarnessX-Seed} the handcrafted HarnessX-style one that we wrote for AppWorld. The evolved rows use the harness each run returns; HarnessX evolves transductively, on test-normal. P@1 and P@2 are pass@1 and pass@2 in \%, averaged over three test runs. API cost is that of the evolution run.}
    \label{tab:qwen9b-appworld-challenge}
    \small
    \setlength{\tabcolsep}{6pt}
    \renewcommand{\arraystretch}{1.08}
    \begin{tabular}{llccc}
      \toprule
      Harness & Mode & P@1 & P@2 & API cost (\$) \\
      \midrule
      Bare             & --           & 0.0  & 0.0 & -- \\
      Bare + format    & --           & 0.0  & 0.0 & -- \\
      \midrule
      HarnessX-Seed    & --           & 21.7 & 32.8 & -- \\
      HarnessX         & Transductive & \textbf{36.0} & \textbf{50.5} & 190.32 \\
      \midrule
      Neutral          & --           & 0.0  & 0.0 & -- \\
      \method{}        & Inductive    & 34.5 & 48.6 & 14.56 \\
      \method{}        & Transductive & 35.2 & 49.2 & 12.51 \\
      \bottomrule
    \end{tabular}
\end{table}

\FloatBarrier
\subsection{Evolution trace on AppWorld}
\label{app:aw-trace}
\begin{figure}[H]
  \centering
  \includegraphics[width=\textwidth]{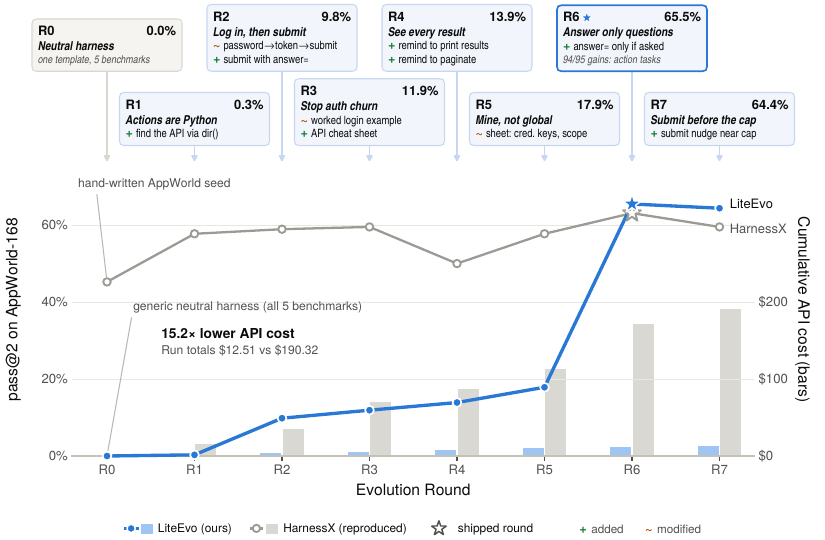}
  \caption{\method{} reaches its best AppWorld harness from a zero-scoring start in six rounds, at 15.2$\times$ lower API cost than HarnessX. One transductive run on AppWorld's 168 test-normal tasks: pass@2 of each round's harness, measured inside the loop (lines); the cumulative API cost of producing each round's harness (bars); and the components each round added ($+$) or modified ($\sim$). The star marks the harness that \method{} returns, whose graded pass@2 over three test runs is 67.7 (Table~\ref{tab:main_results}). Our HarnessX reproduction starts from a hand-written AppWorld seed (Appendix~\ref{app:benchmarks:harnessx}).}
  \label{fig:appworld-evolution}
\end{figure}

Figure~\ref{fig:appworld-evolution} follows the transductive AppWorld run of
Table~\ref{tab:main_results} round by round. Since neither the task text nor
the neutral harness says that actions are Python or how to finish a task, the
agent starts at zero. Each round then adds at most four components and updates
or merges at most two (Table~\ref{tab:qwen9b-library-ops-per-run}), all aimed
at failures in the previous round's trajectories.

\FloatBarrier
\subsection{Transfer across benchmarks}
\label{app:transfer}
We tested whether a library evolved on some benchmarks helps an unseen one, composing harnesses for AppWorld's test-challenge split from a library
built on the other four benchmarks, with no AppWorld component. In three draws
over the 417 tasks, they won 1 of 2{,}502 episodes, where the harnesses that \method{} evolves on
AppWorld itself reach 48.6 and 49.2 pass@2
(Table~\ref{tab:qwen9b-appworld-challenge}) and the neutral harness 0.0. On AgentDiff, a
sibling of AppWorld with 45 tasks, the test is too small to resolve even the
gain of AgentDiff's own library, so it only bounds the transfer. The
composer rarely picks foreign components, and the only kind that transferred is a
loop-breaking processor, which every benchmark's library had invented on its
own. A library's value lies in worked examples written in its environment's own
names and actions, which a new benchmark cannot use. The procedure transfers
instead: a frontier model that reads only the new benchmark's 90 train
transcripts writes worked examples worth 12.6 to 16.3~pp over a harness without
them, 55\% to 71\% of the 23.1~pp that the worked examples of a fully
evolved library add.

\newpage 
\FloatBarrier
\subsection{Analyst setup}
\label{app:analyst}
Table~\ref{tab:qwen9b-analyst-mode} compares failure-oriented group sampling,
in which one analyst reads the grouped trajectories of all workers, with one
analyst per worker, each reading ten failures and three successes of its own
worker. Group sampling costs less in all nine benchmark and regime cells, at
36\% to 75\% of the per-worker cost, and its pass@2 is higher in five cells,
within 0.2~pp in two, and lower in the two ALFWorld cells, by 1.7 and 1.5~pp, where both setups exceed 94.
We use group sampling by default.

\begin{table}[H]
    \centering
    \caption{Analyst setup under Compose (Appendix~\ref{app:refine}; Qwen3.5-9B). \textit{Per-worker} runs one analyst for each worker, on that worker's own trajectories. \textit{GRP} is failure-oriented group sampling: a single analyst reads grouped trajectories of all workers. Scores average three test runs, and API cost is that of one run.}
    \label{tab:qwen9b-analyst-mode}
    \small
    \setlength{\tabcolsep}{3.2pt}
    \renewcommand{\arraystretch}{1.10}
    \begin{tabular}{lllrcccc}
      \toprule
      Benchmark & Mode & Analyst setup & \# Analysts & \shortstack{pass@1\\(\%)} & \shortstack{pass@2\\(\%)} & \shortstack{Best\\Round} & \shortstack{API\\cost (\$)} \\
      \midrule
      \multirow{4}{*}{ALFWorld}
        & \multirow{2}{*}{Transductive}
        & Per-worker & 4 & 96.6 & 98.0 & 7 & 19.26 \\
        & & GRP & 1 & 92.6 & 96.3 & 5 & 8.24 \\
      \cmidrule(lr){2-8}
        & \multirow{2}{*}{Inductive}
        & Per-worker & 4 & 92.5 & 96.3 & 6 & 20.05 \\
        & & GRP & 1 & 90.3 & 94.8 & 4 & 9.17 \\
      \midrule
      \multirow{4}{*}{WebShop}
        & \multirow{2}{*}{Transductive}
        & Per-worker & 4 & 51.8 & 62.7 & 1 & 24.78 \\
        & & GRP & 1 & 54.0 & 62.7 & 5 & 9.00 \\
      \cmidrule(lr){2-8}
        & \multirow{2}{*}{Inductive}
        & Per-worker & 4 & 56.0 & 66.0 & 4 & 21.62 \\
        & & GRP & 1 & 61.0 & 70.0 & 3 & 9.89 \\
      \midrule
      \multirow{4}{*}{AppWorld}
        & \multirow{2}{*}{Transductive}
        & Per-worker & 4 & 40.3 & 54.6 & 3 & 22.78 \\
        & & GRP & 1 & 51.9 & 67.7 & 6 & 12.51 \\
      \cmidrule(lr){2-8}
        & \multirow{2}{*}{Inductive}
        & Per-worker & 4 & 49.4 & 65.3 & 6 & 24.83 \\
        & & GRP & 1 & 49.4 & 65.5 & 7 & 14.56 \\
      \midrule
      \multirow{2}{*}{GAIA}
        & \multirow{2}{*}{Transductive}
        & Per-worker & 4 & 49.8 & 63.4 & 7 & 19.58 \\
        & & GRP & 1 & 53.6 & 65.2 & 7 & 10.59 \\
      \midrule
      \multirow{4}{*}{SWE-bench Verified}
        & \multirow{2}{*}{Transductive}
        & Per-worker & 2 & 54.4 & 61.8 & 1 & 35.92 \\
        & & GRP & 1 & 54.2 & 63.2 & 6 & 24.30 \\
      \cmidrule(lr){2-8}
        & \multirow{2}{*}{Inductive}
        & Per-worker & 2 & 52.7 & 61.8 & 1 & 37.84 \\
        & & GRP & 1 & 58.4 & 66.7 & 4 & 28.50 \\
      \bottomrule
    \end{tabular}
  \end{table}

\newpage 
\FloatBarrier
\subsection{Task selection for the analyst}
\label{app:sampling}
Table~\ref{tab:qwen9b-grp-ablation} changes which task groups the analyst
reads while keeping their number at 13 (7 on SWE-bench Verified). Our default
reads ten mixed groups and three all-fail groups, while the alternatives read
only groups that all workers solved, only groups that no worker solved, or
groups drawn at random, filling a short population from the other groups. The
default gives the best pass@2 on all five benchmarks, though on SWE-bench
Verified only by 0.2~pp over the all-success groups. The choice matters
most on AppWorld, where all-success and all-fail groups reach only 16.9 and
12.3 against 67.7. A failure is plausibly easiest to diagnose next to a success
on the same task, which only mixed groups provide.

\begin{table}[H]
    \centering
    \caption{Which task groups the analyst reads (Qwen3.5-9B, transductive): always 13, or 7 on SWE-bench Verified. \textit{Default} reads 10 mixed-outcome groups and 3 all-fail groups (5 and 2 on SWE-bench Verified). \textit{All-success}, \textit{All-fail} and \textit{Random} read groups that every worker solved, groups that no worker solved, or random groups, each filled up from the other groups when too few exist. Scores average three test runs.}
    \label{tab:qwen9b-grp-ablation}
    \small
    \setlength{\tabcolsep}{5pt}
    \renewcommand{\arraystretch}{1.08}
    \begin{tabular}{llccc}
      \toprule
      Benchmark & Sampling & pass@1 (\%) & pass@2 (\%) & Best Round \\
      \midrule
      \multirow{4}{*}{ALFWorld}
        & Default & \textbf{92.6} & \textbf{96.3} & 5 \\
      \cmidrule(lr){2-5}
        & All-success & 79.9 & 86.6 & 3 \\
        & All-fail & 84.7 & 89.3 & 7 \\
        & Random & 90.1 & 94.3 & 7 \\
      \midrule
      \multirow{4}{*}{WebShop}
        & Default & \textbf{54.0} & \textbf{62.7} & 5 \\
      \cmidrule(lr){2-5}
        & All-success & 49.5 & 62.0 & 7 \\
        & All-fail & 52.7 & 60.7 & 6 \\
        & Random & 52.7 & 62.0 & 1 \\
      \midrule
      \multirow{4}{*}{AppWorld}
        & Default & \textbf{51.9} & \textbf{67.7} & 6 \\
      \cmidrule(lr){2-5}
        & All-success & 13.4 & 16.9 & 5 \\
        & All-fail & 9.3 & 12.3 & 2 \\
        & Random & 43.9 & 57.9 & 6 \\
      \midrule
      \multirow{4}{*}{GAIA}
        & Default & 53.6 & \textbf{65.2} & 7 \\
      \cmidrule(lr){2-5}
        & All-success & \textbf{53.7} & 64.7 & 7 \\
        & All-fail & 46.4 & 57.0 & 4 \\
        & Random & 46.9 & 59.2 & 5 \\
      \midrule
      \multirow{4}{*}{SWE-bench Verified}
        & Default & \textbf{54.2} & \textbf{63.2} & 6 \\
      \cmidrule(lr){2-5}
        & All-success & 53.3 & 63.0 & 2 \\
        & All-fail & \textbf{54.2} & 61.8 & 1 \\
        & Random & 51.4 & 60.0 & 1 \\
      \bottomrule
    \end{tabular}
  \end{table}

\FloatBarrier
\newpage 
\FloatBarrier
\subsection{Library operations per benchmark}
\label{app:library}
\label{app:library-per-run}
Table~\ref{tab:qwen9b-library-ops-per-run} breaks
Table~\ref{tab:qwen9b-library-ops-pooled} down by benchmark: each run adds
components in most rounds and rarely removes one, and in the last round the
composer picks half of the library on SWE-bench Verified (8 of 16 components)
and all of it on AppWorld (13 of 13).

\providecommand{\method}{LiteEvo}
\begin{table}[H]
    \centering
    \caption{Per-benchmark breakdown of Table~\ref{tab:qwen9b-library-ops-pooled}. \textit{Picked / Size} is the number of components the composer selects over the number it can select from. GAIA counts are reconstructed from library snapshots.}
    \label{tab:qwen9b-library-ops-per-run}
    \small
    \setlength{\tabcolsep}{4.5pt}
    \renewcommand{\arraystretch}{1.08}
    \begin{tabular}{@{}llccccccccc@{}}
      \toprule
      Benchmark & Round & 0 & 1 & 2 & 3 & 4 & 5 & 6 & 7 & Total \\
      \midrule
      \multirow{5}{*}{ALFWorld}
        & Add           & 4   & 4   & 3     & 2     & 1     & 2     & 1     & 2     & 19 \\
        & Update        & 0   & 0   & 0     & 0     & 1     & 0     & 0     & 1     & 2  \\
        & Merge         & 0   & 0   & 1     & 1     & 0     & 0     & 0     & 0     & 2  \\
        & Drop          & 0   & 0   & 1     & 0     & 0     & 0     & 0     & 0     & 1  \\
        & Picked / Size & 4/4 & 8/8 & 10/10 & 12/12 & 13/13 & 15/15 & 14/16 & 16/18 & -- \\
      \midrule
      \multirow{5}{*}{WebShop}
        & Add           & 3   & 2   & 2   & 1   & 1   & 3     & 2     & 0    & 14 \\
        & Update        & 0   & 0   & 0   & 0   & 0   & 0     & 0     & 3    & 3  \\
        & Merge         & 0   & 0   & 0   & 0   & 0   & 0     & 0     & 0    & 0  \\
        & Drop          & 0   & 0   & 0   & 0   & 0   & 0     & 0     & 0    & 0  \\
        & Picked / Size & 3/3 & 5/5 & 7/7 & 8/8 & 9/9 & 12/12 & 11/14 & 9/14 & -- \\
      \midrule
      \multirow{5}{*}{AppWorld}
        & Add           & 4   & 2   & 1   & 2   & 0   & 2    & 2     & 2     & 15 \\
        & Update        & 0   & 1   & 0   & 1   & 1   & 0    & 0     & 1     & 4  \\
        & Merge         & 0   & 1   & 2   & 0   & 0   & 0    & 0     & 0     & 3  \\
        & Drop          & 0   & 0   & 0   & 0   & 0   & 0    & 0     & 1     & 1  \\
        & Picked / Size & 4/4 & 5/5 & 6/6 & 7/8 & 7/8 & 9/10 & 11/12 & 13/13 & -- \\
      \midrule
      \multirow{5}{*}{GAIA}
        & Add           & 4   & 3   & 4     & 3     & 3     & 1     & 1     & 2     & 21 \\
        & Update        & 0   & 0   & 2     & 1     & 1     & 0     & 1     & 1     & 6  \\
        & Merge         & 0   & 0   & 0     & 0     & 0     & 1     & 0     & 0     & 1  \\
        & Drop          & 0   & 0   & 0     & 0     & 0     & 0     & 0     & 0     & 0  \\
        & Picked / Size & 4/4 & 7/7 & 10/11 & 13/14 & 14/17 & 13/18 & 14/19 & 16/21 & -- \\
      \midrule
      \multirow{5}{*}{\shortstack[l]{SWE-bench\\Verified}}
        & Add           & 4   & 0   & 2   & 2   & 1   & 3    & 2    & 2    & 16 \\
        & Update        & 0   & 3   & 2   & 0   & 2   & 0    & 2    & 1    & 10 \\
        & Merge         & 0   & 0   & 0   & 1   & 0   & 0    & 0    & 0    & 1  \\
        & Drop          & 0   & 0   & 0   & 0   & 0   & 0    & 0    & 0    & 0  \\
        & Picked / Size & 4/4 & 4/4 & 6/6 & 8/8 & 6/9 & 8/12 & 6/14 & 8/16 & -- \\
      \bottomrule
    \end{tabular}
\end{table}

\newpage 
\FloatBarrier
\subsection{Memory in evolved harnesses}
\label{app:memory}
The component kinds of \method{} include no memory that persists across
episodes, yet the evolved harnesses keep information in two ways. Within an
episode, processors can use the scratchpad and the step history: one GAIA
processor collects the URLs that returned an access wall and warns more
strongly as the list grows, and another quotes back any search that the agent
repeats verbatim. Across episodes, knowledge and prompt fragments carry
what earlier rounds observed, as when the AppWorld harness spells out the whole
login procedure that its early rounds had to discover; no state passes from
one episode to the next at run time.

Table~\ref{tab:qwen9b-memory-ablation} tests whether an explicit memory across
episodes helps, with each variant a hand-written component pinned into every
composition. The action book stores a failed action with the next action that
worked, the URL blacklist stores domains that blocked access in at least two
tasks, and login recipes store the API calls that logged in. Every variant
scores lower than the default harness, which reaches 65.2 on GAIA against 55.8
and 58.4, and 67.7 on AppWorld against 15.4. The comparison is not controlled: each variant was measured once, and the
AppWorld variants also ran with a different code version and load than the
default. The action book, which we had registered in advance as a weak variant,
stopped growing at 19 distinct entries by round~2, and none of the variants
shows a gain.

\begin{table}[H]
    \centering
    \caption{Pinning into every composition a hand-written memory that persists across episodes (Qwen3.5-9B, transductive). The memory reads and writes a store shared by episodes, and no episode reads entries from attempts on its own task. \textit{Action book} stores a failed action with the next action that worked. \textit{URL blacklist} stores, as an avoid-list, web domains that blocked access in at least two tasks. \textit{Login recipes} stores the API calls that logged in. \textit{None} rows are the default runs of Table~\ref{tab:qwen9b-grp-ablation}.}
    \label{tab:qwen9b-memory-ablation}
    \small
    \setlength{\tabcolsep}{5pt}
    \renewcommand{\arraystretch}{1.08}
    \begin{tabular}{llccc}
      \toprule
      Benchmark & Enforced Memory & pass@1 (\%) & pass@2 (\%) & Best Round \\
      \midrule
      \multirow{3}{*}{GAIA}
        & None (default) & \textbf{53.6} & \textbf{65.2} & 7 \\
      \cmidrule(lr){2-5}
        & Action book & 45.1 & 55.8 & 5 \\
        & URL blacklist & 46.1 & 58.4 & 7 \\
      \midrule
      \multirow{2}{*}{AppWorld}
        & None (default) & \textbf{51.9} & \textbf{67.7} & 6 \\
      \cmidrule(lr){2-5}
        & Login recipes & 13.2 & 15.4 & 6 \\
      \bottomrule
    \end{tabular}
\end{table}

\newpage 
\FloatBarrier
\subsection{Merging the editor and the librarian}
\label{app:merge}
Table~\ref{tab:qwen9b-merge-ablation} merges the editor and the librarian into
one meta-agent, which also reads the replay results of all candidate processors.
The merged runs score lower on all four benchmarks, most on AppWorld, where
inductive pass@2 falls from 65.3 to 14.0. These runs, however, differ from the
defaults in more than the merge, and no split control arm was run beside them;
the only controlled comparison we have, a pilot on WebShop, found no
difference. On GAIA, the merged run used one analyst per worker,
ran without a working code tool, and was measured once. We keep the two meta-agents separate.

\begin{table}[H]
    \centering
    \caption{Merging the editor and the librarian into one meta-agent (Qwen3.5-9B). \textit{None} keeps the two meta-agents separate: the editor selects and rewrites the analyst's candidates, and the librarian applies them to the library. \textit{Editor + Librarian} merges them and uses one analyst per worker. The None rows are the per-worker runs of Table~\ref{tab:qwen9b-analyst-mode}, except on GAIA, whose default is the group-sampling run of Table~\ref{tab:qwen9b-grp-ablation}. GAIA is transductive; the other benchmarks are inductive and scored on held-out tasks.}
    \label{tab:qwen9b-merge-ablation}
    \small
    \setlength{\tabcolsep}{5pt}
    \renewcommand{\arraystretch}{1.08}
    \begin{tabular}{llcc}
      \toprule
      Benchmark & Merge & pass@1 (\%) & pass@2 (\%) \\
      \midrule
      \multirow{2}{*}{GAIA}
        & None (default) & \textbf{53.6} & \textbf{65.2} \\
      \cmidrule(lr){2-4}
        & Editor + Librarian & 48.1 & 59.9 \\
      \midrule
      \multirow{2}{*}{ALFWorld}
        & None (default) & \textbf{92.5} & \textbf{96.3} \\
      \cmidrule(lr){2-4}
        & Editor + Librarian & 91.0 & 92.5 \\
      \midrule
      \multirow{2}{*}{WebShop}
        & None (default) & \textbf{56.0} & \textbf{66.0} \\
      \cmidrule(lr){2-4}
        & Editor + Librarian & 52.5 & 61.0 \\
      \midrule
      \multirow{2}{*}{AppWorld}
        & None (default) & \textbf{49.4} & \textbf{65.3} \\
      \cmidrule(lr){2-4}
        & Editor + Librarian & 10.6 & 14.0 \\
      \bottomrule
    \end{tabular}
\end{table}

\FloatBarrier
\subsection{Refine or compose}
\label{app:refine}
Table~\ref{tab:qwen9b-refine-compose} compares two ways for the composer to
produce a harness: Compose, our default, builds each harness afresh from the
library, whereas Refine composes once after the first round and then edits the
carried harness in every round, with one analyst per worker in both.
Compose scores higher on average in both regimes, with 68.1 against 64.3
pass@2 transductively and 72.4 against 65.8 inductively, while Refine wins on
WebShop in the transductive regime and on SWE-bench Verified in the inductive
one.

\providecommand{\method}{LiteEvo}
\begin{table}[H]
    \centering
    \caption{Two ways for the composer to produce a harness (Qwen3.5-9B). \textit{Refine} composes a harness after the first round and then edits it in every round; \textit{Compose} builds a fresh harness from the library in every round. Both use one analyst per worker. P@1 and P@2 are pass@1 and pass@2 in \%, averaged over three test runs. Bold marks the better mode within each regime. \textit{Avg.} averages five benchmarks transductively and four inductively, since GAIA has no inductive run.}
    \label{tab:qwen9b-refine-compose}
    \small
    \setlength{\tabcolsep}{2.5pt}
    \renewcommand{\arraystretch}{1.08}
    \begin{tabular}{@{}llcccccccccccc@{}}
      \toprule
      & & \multicolumn{2}{c}{ALFWorld} & \multicolumn{2}{c}{WebShop} & \multicolumn{2}{c}{AppWorld}
      & \multicolumn{2}{c}{GAIA} & \multicolumn{2}{c}{\shortstack{SWE-bench\\Verified}} & \multicolumn{2}{c}{Avg.} \\
      \cmidrule(lr){3-4} \cmidrule(lr){5-6} \cmidrule(lr){7-8} \cmidrule(lr){9-10} \cmidrule(lr){11-12} \cmidrule(lr){13-14}
      Regime & Mode & P@1 & P@2 & P@1 & P@2 & P@1 & P@2 & P@1 & P@2 & P@1 & P@2 & P@1 & P@2 \\
      \midrule
      \multirow{2}{*}{Transductive}
        & Refine  & 88.3 & 93.9 & \textbf{58.6} & \textbf{67.8} & 27.4 & 41.6 & 48.0 & 58.8 & 50.7 & 59.4 & 54.6 & 64.3 \\
        & Compose & \textbf{96.6} & \textbf{98.0} & 51.8 & 62.7 & \textbf{40.3} & \textbf{54.6}
                  & \textbf{49.8} & \textbf{63.4} & \textbf{54.4} & \textbf{61.8} & \textbf{58.6} & \textbf{68.1} \\
      \midrule
      \multirow{2}{*}{Inductive}
        & Refine  & \textbf{93.1} & 95.5 & 52.0 & 59.2 & 31.4 & 45.6 & -- & -- & \textbf{53.9} & \textbf{63.1} & 57.6 & 65.8 \\
        & Compose & 92.5 & \textbf{96.3} & \textbf{56.0} & \textbf{66.0} & \textbf{49.4} & \textbf{65.3} & -- & -- & 52.7 & 61.8 & \textbf{62.7} & \textbf{72.4} \\
      \bottomrule
    \end{tabular}
\end{table}

\FloatBarrier

\clearpage
\section{Limitations}
\label{app:limitations}
\method{} fits one library per benchmark: its harnesses transfer to unseen
tasks of the same benchmark and, on AppWorld, to unseen apps, but a library
evolved on other benchmarks did not help an unseen one
(Appendix~\ref{app:transfer}). The procedure, however, carries across
benchmarks: the same loop builds a working library for each benchmark from the
same neutral start. We therefore see \method{} as a step towards harness
evolution that is fully automated and generalizes across benchmarks. On
AppWorld's test-challenge split, HarnessX's harness scores higher than ours
(Appendix~\ref{app:aw-challenge}).

Each result comes from a single evolution run, so the tables do not show how
much the loop's outcome varies from run to run. On GAIA, the agent can still
reach answers through copies of the benchmark that the web tools do not block:
in round~7, which ran our returned harness, 10 winning episodes reached
such a copy before answering. Scoring them as failures would leave that round's
pass@2 no lower than 62.9, against 64.6 as measured, a bound that covers only
the loop's episodes (Appendix~\ref{app:benchmarks:gaia}). One evolved component
also points the agent at dataset mirror pages.

The librarian mostly adds components, and we did not try more active
management of the library. For lack of resources, we did not use a flagship
model as the agent or test the hardest agentic benchmarks; tasks outside
agentic benchmarks are out of scope. We wrote the meta-agents' instructions
once and never optimized them; letting a frontier model design these
instructions, packaging \method{} as a skill for coding agents such as Claude
Code, and combining it with fine-tuning or reinforcement learning are left to
future work.

\end{document}